%% file: main.tex
\documentclass[]{interact}

\usepackage{epstopdf}% To incorporate .eps illustrations using PDFLaTeX, etc.
\usepackage{subfigure}

\usepackage{color}
\def\revised_color{red}
\usepackage[markup=default, defaultcolor=red]{changes}
\usepackage{cancel}

\usepackage[numbers,sort&compress]{natbib}% Citation support using natbib.sty
\bibpunct[, ]{[}{]}{,}{n}{,}{,}% Citation support using natbib.sty
\renewcommand\bibfont{\fontsize{10}{12}\selectfont}% Bibliography support using natbib.sty
\makeatletter% @ becomes a letter
\def\NAT@def@citea{\def\@citea{\NAT@separator}}% Suppress spaces between citations using natbib.sty
\makeatother% @ becomes a symbol again

\newcommand{\figref}[1]{{Fig. \ref{#1}}}

\theoremstyle{plain}% Theorem-like structures provided by amsthm.sty

\theoremstyle{definition}

\theoremstyle{remark}

\newcommand{\pdif}[2]{\dfrac{\partial #1}{\partial #2}}
\renewcommand{\eqref}[1]{{Eq.} (\ref{#1})}

\begin{document}

%\articletype{ARTICLE TEMPLATE}% Specify the article type or omit as appropriate

\title{Dynamics Computation of Soft-Rigid Hybrid-Link System and Its Application to Motion Analysis of an Athlete Wearing Sport Prosthesis}

\author{
\name{Sunghee Kim\textsuperscript{a}, 
Yuta Shimane\textsuperscript{a},
Taiki Ishigaki\textsuperscript{b},
and Ko Yamamoto\textsuperscript{a}\thanks{CONTACT Ko Yamamoto. Email: yamamoto.ko@ynl.t.u-tokyo.ac.jp. This is an original manuscript of an article published by Taylor \& Francis in Advanced Robotics, Vol.40, No.4, 2026 on 12 Jan 2026, available online: https://doi.org/10.1080/01691864.2025.2611426.}
}
\affil{\textsuperscript{a}Department of Mechano-Informatics, University of Tokyo, 7-3-1 Hongo, Bunkyo-Ku, Tokyo, Japan
; \textsuperscript{b}Research Institute for Science and Technology, Tokyo University of Science, Katsushika-ku, Tokyo, Japan
}
}

% \author{
% \name{Sunghee Kim\textsuperscript{a}, 
% Yuta Shimane\textsuperscript{a},
% Taiki Ishigaki\textsuperscript{b},
% and Ko Yamamoto\textsuperscript{a}\thanks{CONTACT Ko Yamamoto. Email: yamamoto.ko@ynl.t.u-tokyo.ac.jp}}
% \affil{\textsuperscript{a}Department of Mechano-Informatics, University of Tokyo, 7-3-1 Hongo, Bunkyo-Ku, Tokyo, Japan
% ; \textsuperscript{b}Research Institute for Science and Technology, Tokyo University of Science, Katsushika-ku, Tokyo, Japan
% }
% }

\maketitle

\begin{abstract}
This paper presents a motion analysis framework for an athlete wearing sport-specific flexible prosthesis based on the soft-rigid hybrid-link system.
Such a motion analysis is a challenging problem because we need to consider the interaction force between the rigid human skeleton system and a flexible prosthesis.
However, most of human musculoskeletal models are based on the computation framework of a rigid-body multi-link system.
Recently in soft robotics research field, fast and efficient modeling methods were developed for a flexible rod deformation, which allows us to build a hybrid-link system that integrates rigid-link and soft-bodies in a unified formulation.
We apply inverse kinematics of the hybrid-link system to motion reconstruction from a motion captured data, and also present the estimation of the joint torques and ground reaction force by inverse dynamics. 
Through a human subject experiment, we show that the inverse dynamics achieved approximately 12\% error on the ground reaction force estimation.
Furthermore, we provide the muscle force estimation considering muscle amputation and interaction force with the prosthesis leg deformation.
\end{abstract}

\begin{keywords}
Kinematics and Dynamics; Biomechanics; Soft robotics
\end{keywords}

%%%%%%%%%%%%%%%%%%%%%%%%%%%%%%%%%%%%%%%%%%%%%%%%%%%%%%%%%%%%%%%%%%%%%%
\input{contents/introduction}
\input{contents/methods}
\input{contents/results}
\input{contents/discussions}
\input{contents/conclusion}
%%%%%%%%%%%%%%%%%%%%%%%%%%%%%%%%%%%%%%%%%%%%%%%%%%%%%%%%%%%%%%%%%%%%%%

\appendix
\input{contents/apdx_msforce_calc}
\input{contents/apdx_hybrid_link_inertia}

\section*{Acknowledgements}
The experiment of this research was conducted thanks to the cooperation of Dr. Sayaka Fujiwara, the Faculty of Medicine, University of Tokyo.
The authors also thank Yosuke Ikegami and Kazuya Tomabechi for their advise on the musculo-skeletal model and motion capture measurement, Kensho Hiraoka for his support for the motion capture measurement, and Akihiro Sakurai for his support for the visualization of muscle activities.

\section*{Disclosure statement}
No potential conflict of interest was reported by the authors.

\section*{Funding}
This work was supported by JSPS KAKENHI 21H01282.

\if0
\section*{Notes on contributor(s)}

An unnumbered section, e.g.\ \verb"\section*{Notes on contributors}", may be included \emph{in the non-anonymous version} if required. A photograph may be added if requested.

\section*{Nomenclature/Notation}

An unnumbered section, e.g.\ \verb"\section*{Nomenclature}" (or \verb"\section*{Notation}"), may be included if required, before any Notes or References.

\section*{Notes}

An unnumbered `Notes' section may be included before the References (if using the \verb"endnotes" package, use the command \verb"\theendnotes" where the notes are to appear, instead of creating a \verb"\section*").
\fi
% \bibliographystyle{tfnlm}
%\bibliography{bib/ar_hydracer}

\appendix

\bibliographystyle{tfnlm} 
\bibliography{reference}

\end{document}

%% file: contents/introduction.tex
\section{Introduction}
A prosthesis leg specialized for sports opens the door for individuals with congenital or acquired limb disabilities to enjoy sports and participate in competitions such as the Paralympics \cite{Hobara2015}.
As the developments in robotics and biomechanics, motion analysis based on a human musculo-skeletal model \cite{nakamura2005somatosensory, delp2007opensim} will provide a quantitative knowledge on performance analysis, training, and injury prevention not only for able-bodied athlete but also for Para-athlete.
However, motion analysis of an athlete wearing a flexible prosthesis is a challenging problem because we need to consider the interaction force between the rigid human skeleton system and a flexible prosthesis.
Most of human musculoskeletal models are based on kinematics and dynamics computation framework of a rigid-body multi-link system and cannot directly deal with a continuous deformation of a leaf-spring prosthesis leg. 
Previous studies employed a rigid-body model \cite{rigney2016prosthesis} or simple spring-damper models \cite{kleesattel2018inverse,murai2018can} for a prosthesis sprinting motion.
These models were too simple to appropriately model the deformation of a prosthesis.

\if0
Human motion analysis encompasses two essential aspects: kinematics and kinetics.
Kinetic analysis explores the forces influencing a system and generating dynamic motions. 
Previous studies have applied inverse dynamics based on robotics principles to investigate motion kinetics \cite{delp2007opensim}, \cite{nakamura2005somatosensory}, \cite{samy2019real}. 
These studies represented the human model as a musculoskeletal model with a rigid multi-link system.
However, analyzing interactions between humans and athletic tools having flexible deformations challenges with this method.  
For instance, while earlier studies effectively analyzed motion with running-specific prostheses \cite{bruggemann2008biomechanics}, \cite{murai2018can}, \cite{kleesattel2018inverse}, \cite{rigney2016prosthesis}, they represented the prosthesis using spring-damper models with rigid segments, limiting the accurate capture of detailed deformations.
\fi

The field of soft robotics has explored various methods for modeling flexible deformation. 
The finite element method (FEM) is a representative technique to calculate a continuous deformation of a soft structure \cite{duriez2013control,mendhurwar2020system,kays2017effect,vivaldi2018study}. 
However, one of the known limitations in FEM is its high computational cost \cite{runge2017framework,schegg2022review}, especially in the case that a complicated shape is divided into a lot of meshes. 
While FEM is suitable for an object with complicated shape, there are methods specific for a rod or beam structure based on the Cosserat rod theory \cite{Atman2005nonlinear}, which can offer much faster computation and is suitable for a leaf-spring prosthesis. 
This theory conceptualizes a rod as a deformable vector with an orientation, allowing each point along its central axis coordinate to be described using a local frame.
Thanks to this characteristic, the analysis of a soft continuum based on the Cosserat rod theory shows a framework similar to that of the rigid-body multi-link system.
In particular, the Piecewise Constant Strain (PCS) model \cite{renda2018discrete}, \cite{renda2018geometric} is compatible with the rigid-body multi-link system because it discretizes a flexible rod or beam structure into a finite number of segments and assumes a constant strain in each segment. 
Because of its compatibility, some studies have proposed hybrid-link models that integrated rigid-links and soft-bodies \cite{ishigaki2021dynamics,Mathew2023}.

\begin{figure}[t]
 \centering
% \begin{subfigure}{\linewidth}
     \centering
    \includegraphics[width=1.0\linewidth]{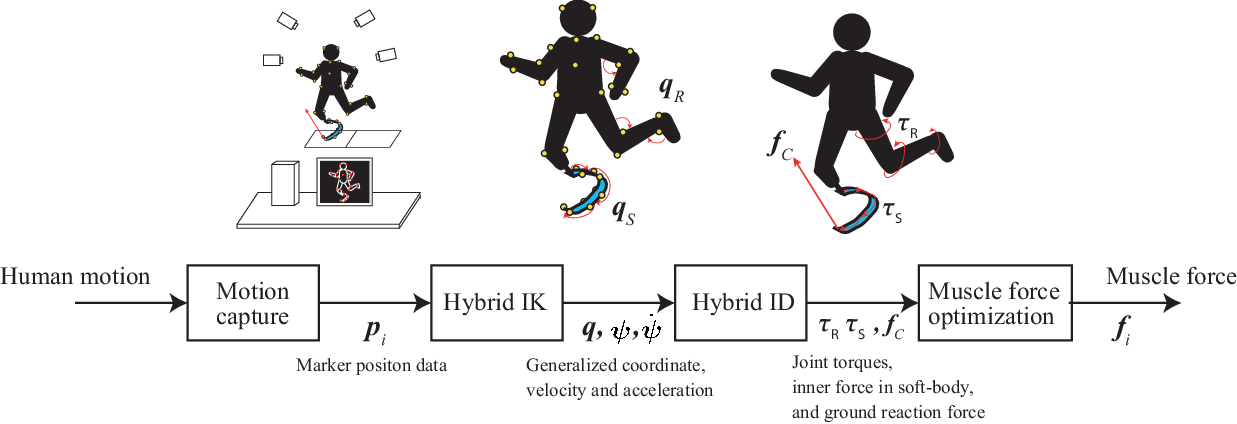}
%  \caption{}
%\end{subfigure}
%\begin{subfigure}{\linewidth}
%  \centering
%  \includegraphics[width=.7\linewidth]{figure/id_result_intro.png}
%  \caption{}
%\end{subfigure}
  \caption{Overview for the motion analysis framework based on the hybrid-link system. From a motion capture system, we reconstruct human motion and prosthesis deformation by inverse kinematics. Then, inverse dynamics calculates the joint torques, inner force in the prosthesis and the ground reaction force. From the obtained joint torques and interaction force from the prosthesis, we estimate muscle tendon forces by an optimization.}
  \label{fig:figure_intro}
  \vspace{-0.2cm}
\end{figure}

In this study, we apply the framework of kinematics and dynamics computation of a hybrid-link system to dynamic motion analysis of an athlete wearing a sports-specific prosthesis.
\figref{fig:figure_intro} provides an overview of the motion analysis framework, including inverse kinematics (IK) calculation for the motion capture measurement, inverse dynamics (ID) calculation for the estimation of joint torques and ground reaction forces, and muscle-force optimization considering the interaction force between the human-body and prosthesis.
Through a human subject experiment, we validate that dynamics computation based on the hybrid-link can provide a good estimation of the ground reaction force.
Moreover, we demonstrate a visualization of the muscle-force optimization result during the motion.

The rest of this paper is organized as follows.
Section \ref{sect:methods} describes the basic formulations of the PCS model and the hybrid-link system, and presents the inverse kinematics of the hybrid-link system for a motion capture measurement.
Then, the inverse dynamics of the hybrid-link system for the estimation of the ground reaction force, and muscle force optimization are introduced.
Section \ref{sect:results} reports experimental results including motion capture measurement, ground force estimation by inverse dynamics, and a visualization of the muscle force estimation, providing a preliminary result on the muscle force estimation.
Finally, Section \ref{sect:discussions} provides discussions on the experimental results, and Section \ref{section:conclusion} summarizes the obtained results and concludes the paper.

A part of this paper was reported in our previous conference paper \cite{kim2022inverse}.
This study provides more dynamical analyses using inverse dynamics and muscle-force optimization whereas \cite{kim2022inverse} reported a simple analysis of the inverse kinematics for the motion capture measurement, mainly presented in Section \ref{sect:hybrid_ik}.

Ethics approval for the experimental procedures in this study was granted by the Human Research Committee at Graduation School of Information Science and Technology, the University of Tokyo (approval no.: UT-IST-RE-220209).
Hereafter, we use the following basic notations and definitions.
\begin{itemize}
   \item $\bm{E}$ and $\bm{O}$ denote the unit and zero matrices, respectively.
   \item $SO(3)$ is the three dimensional special orthogonal group, and $so(3)$ is its Lie algebra.
   \item $SE(3)$ is the three dimensional special Euclid group, and $se(3)$ is its Lie algebra.
%   \item $\bm{p} \in \mathbb{R}^3$ denotes a position vector.
%   \item $\bm{R} \in SO(3)$ denotes an orientation matrix.
   \item Given a vector $\bm{a} = \begin{bmatrix} a_1 & a_2 & a_3 \end{bmatrix}^T \in \mathbb{R}^3$, the skew-symmetric matrix $[\bm{a} \times]$ is defined as 
       \begin{align}
        [\bm{a} \times] := 
        \begin{bmatrix}
        0 & -a_3 & a_2 \\
        a_3 & 0 & -a_1\\
        -a_2 & a_1 & 0
        \end{bmatrix} 
        \in so(3).
    \end{align}
    \item Given three-dimensional vectors $\bm{a}$ and $\bm{b}$, and a six-dimensional vector $\bm{c} = [\bm{a}^T \ \bm{b}^T]^T$,  operators $[ \bm{c} \times]$ and $[ \bm{c}_\bullet]$ are defined as
    \begin{align}
        [ \bm{c} \times ] &:=
        \begin{bmatrix}
            [\bm{a} \times] & \bm{b} \\
            \bm{0}^T & 0
        \end{bmatrix}
        \in se(3),
        \\
        [ \bm{c}{}_\bullet] &:=
        \begin{bmatrix}
            [\bm{a} \times] & \bm{O} \\
            [\bm{b} \times] & [\bm{a} \times]
        \end{bmatrix}
        .
    \end{align}
    \item $\| \bm{x} \|_{\bm{W}}^2 = \bm{x}^T \bm{W} \bm{x}$ is a weighted squared norm of a vector $\bm{x}$ with a weighting matrix $\bm{W}$. 
\end{itemize}

%% file: contents/methods.tex
\begin{figure}[t]
  \centering
  \includegraphics[width=.55\linewidth]{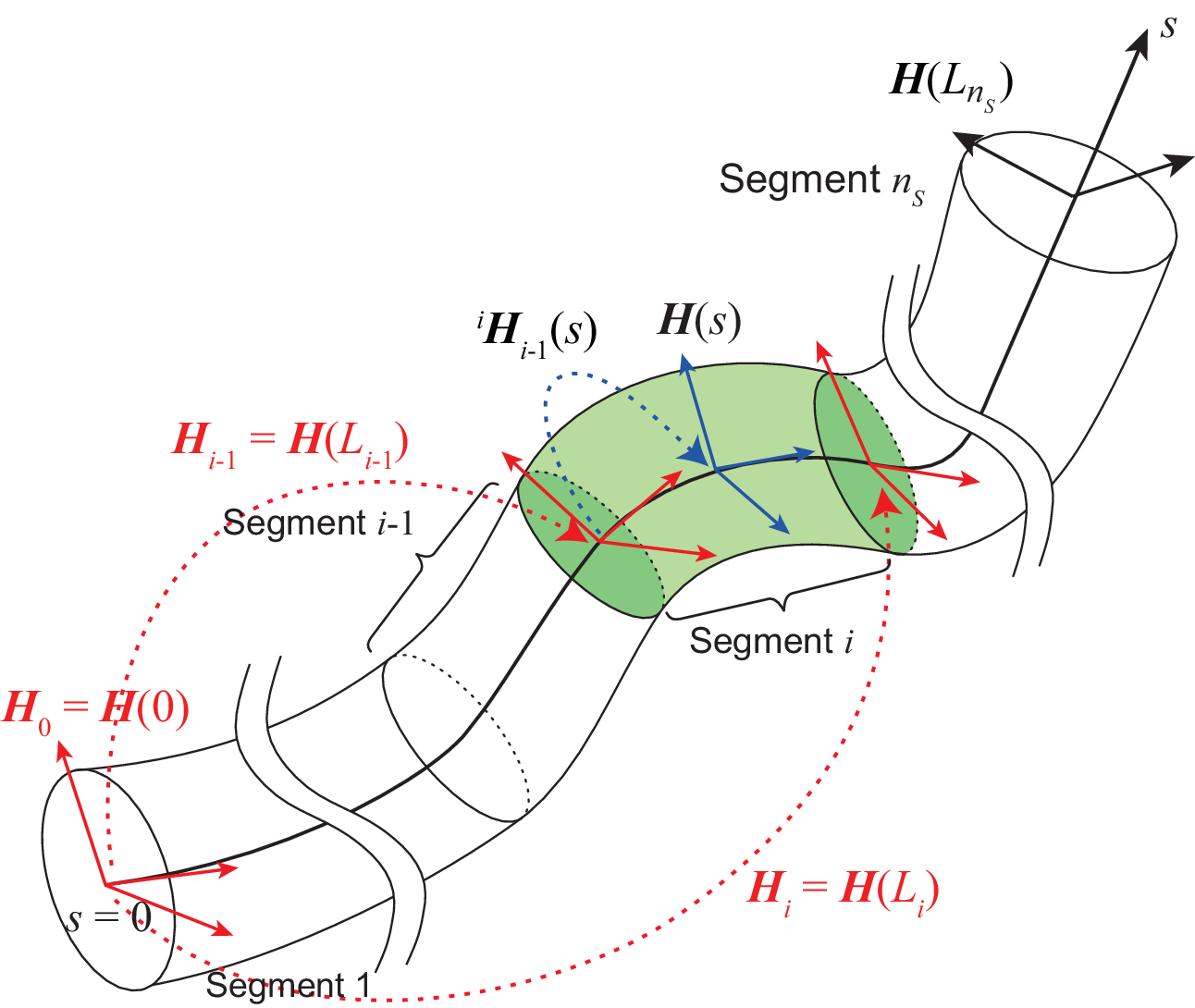}
  \caption{Schematic view of the PCS model }
  \label{fig:pcs_schematics}
\end{figure}

\section{Methods\label{sect:methods}}
\subsection{PCS Model }
\subsubsection{Kinematics of PCS Model}
In Cosserat rod theory, the microsolid of a soft continuum along the central axis coordinate $s \in \mathbb{R}$ at a certain time $t \in \mathbb{R}$ is represented by a configuration curve $\bm{H}(s,t) \in SE(3)$ as
\begin{align}
    \bm{H}(s,t)
    =
    % \bm{H}(\bm{R}(s,t),\bm{p}(s,t)),
    \begin{bmatrix}
    \bm{R}(s,t) & \bm{p}(s,t) \\
    \bm{0}^T & 1
    \end{bmatrix}
\end{align}
where $\bm{R}(s,t) \in SO(3)$ is the rotational matrix and $\bm{p}(s,t) \in \mathbb{R}^3$ is the position vector.
Hereinafter, $t$ or $s$ is omitted for simplicity when its meaning is obvious.
Given the configuration curve $\bm{H}(s)$, a six-dimensional displacement $\bm{\xi}(s) \in \mathbb{R}^6$ is defined with the operator $[\bm{\xi}(s) \times]$ as
\begin{align}
\label{eq:xi_def}
    [\bm{\xi}(s) \times] 
    :=
    \bm{H}(s)^{-1} %\bm{H}^{\prime}(s)
    \dfrac{\partial \bm{H}}{\partial s} .
%    = \begin{bmatrix}
%    [\bm{k}(s) \times ] & \bm{u}(s) \\
%    \bm{0}^T & 0
%    \end{bmatrix}
\end{align}
%where $\bm{u}(s) \in \mathbb{R}^3$ and $\bm{k}(s) \in \mathbb{R}^3$ denote the linear and angular strain, respectively.
In this paper, we refer to the continuous deformation defined by \eqref{eq:xi_def} as {\it strain} although it is not exactly same with the strict definition in material mechanics.

\figref{fig:pcs_schematics} shows a schematic view of the PCS model \cite{renda2018discrete},
in which a rod is divided into a finite number of segments.
Let $n_s$ denote the number of segments, and define $L_{i-1} \le s \le L_{i}$ as segment $i \ (i = 1, \cdots, n_s)$ with $L_0 = 0$.
The PCS model assumes that each segment has a constant strain
$\bm{\xi}_i$, i.e.,
\begin{align}
    \label{eq:constant_strain}
    \bm{\xi}_i := \bm{\xi}(s)
%    = \begin{bmatrix}
%        {\bm{k}_i} \\ {\bm{u}_i}
%    \end{bmatrix}.
    \quad (L_{i-1} \le s \le L_i) .
\end{align}
Substituting \eqref{eq:constant_strain} into \eqref{eq:xi_def}, we obtain a first-order differential equation of $\bm{H}(s)$ as
\begin{align}
    \pdif{\bm{H}}{s} = \bm{H}(s) [ \bm{\xi}_i \times ] .
\end{align}
From the solution of this equation, we can calculate the configuration curve in segment $i$ as
\begin{align}
    \label{eq:pcs_fwd_kinematics0}
    \bm{H}_i(s) = \bm{H}(L_{i-1}) \exp\{ (s-L_{i-1}) [\bm{\xi}_i \times] \} .
\end{align}
Note that we add the index $i$ in the right-lower of $\bm{H}(s)$ to emphasize that this equation is satisfied in segment $i$.
Then, we define
\begin{align}
    \bm{H}_{i-1} &:= \bm{H}(L_{i-1}), \ \mbox{and}
    \\
    \label{eq:xi_diff_solution}
    {}^{i-1}\bm{H}_i(s, \bm{\xi}_i) &:= \exp\{ (s-L_{i-1}) [\bm{\xi}_i \times]\},
\end{align}
\eqref{eq:pcs_fwd_kinematics0} can be rewritten as
\begin{align}
    \label{eq:pcs_fwd_kinematics}
    \bm{H}_i(s) = \bm{H}_{i-1} {}^{i-1}\bm{H}_i(s, \bm{\xi}_i).
\end{align}
Moreover, using a notation of
\begin{align}
    \bm{H}_i = \bm{H}_i(L_i), \ \mbox{and}
    \
    {}^{i-1}\bm{H}_i(L_i, \bm{\xi}_i) = {}^{i-1} \bm{H}_i(\bm{\xi}_i)
\end{align}
we can obtain the following recursion that is similar to the traditional rigid-link system.
\begin{align}
    \label{eq:pcs_fwd_kinematics1}
    \bm{H}_i = \bm{H}_{i-1} {}^{i-1}\bm{H}_i(\bm{\xi}_i) .
\end{align}

Given $\bm{\xi}_i$ of all segments, we can calculate the configuration curve at any coordinate by recursively applying \eqref{eq:pcs_fwd_kinematics} or \eqref{eq:pcs_fwd_kinematics1}.
While \eqref{eq:pcs_fwd_kinematics} represents the forward kinematics inside segment $i$, \eqref{eq:pcs_fwd_kinematics1} represents that for the beginning to the end of segment $i$.
Therefore, we can define the generalized coordinate of the PCS model as
\begin{align}
    \label{eq:pcs_gen_coord}
    \bm{q}_s
    =
    \begin{bmatrix}
        \bm{\xi}_1^T & \cdots & \bm{\xi}_{n_s}^T
    \end{bmatrix}^T \in \mathbb{R}^{6n_s}.
\end{align}
If the base of segment 1 ($s = 0$) is not fixed to the environment, we can add $\bm{H}_0$ in the generalized coordinate $\bm{q}_s$.

In \eqref{eq:xi_def}, replacing $s$ with $t$ yields the definition of the spatial velocity \cite{Featherstone2008} $\bm{\eta}(s)$ as
\begin{align}
    [\bm{\eta}(s) \times] := \bm{H}(s)^{-1} \pdif{\bm{H}}{t}
\end{align}
Let $\bm{\eta}_i(s)$ denote the spatial velocity at the coordinate $s$ in segment $i$. 
From the definition of $\bm{\eta}_i(s)$ and  $\bm{\xi}(s)$, the following differential equation can be derived: 
\begin{align} 
    \label{eq:nu(s)_rewrite}
    \pdif{\bm{\eta}_i}{s}
    =
    \dot{\bm{\xi}}_i -  [\bm{\xi}_i {}_\bullet ]\bm{\eta}(s) .
\end{align}
The solution of \eqref{eq:nu(s)_rewrite} is obtained as follows:
\begin{align}
    \label{eq:eta}
    \bm{\eta}_i (s) &= 
    {}^i\bm{A}(s)^{-1}
    \{ \bm{\eta} (L_{i-1}) + {}^i\bm{T}(s) \dot{\bm{\xi}}_i \},
    \\
    & {}^i\bm{A}(s) := \exp \{ (s - L_{i-1} ) [\bm{\xi}_i {}_\bullet ] \},
%    &= 
%    \begin{bmatrix}
%        {}^{i-1}\bm{R}_i(s)^T& \bm{O} \\ 
%        -{}^{i-1}\bm{R}_i(s)^T\left[{}^{i-1}\bm{p}_i(s) \times\right] & {}^{i-1}\bm{R}_i(s)^T
%    \end{bmatrix}
    \\
    & {}^i\bm{T}(s) := \int_{L_{i-1}}^s \exp\{(x-L_{i-1}) [\bm{\xi}_i {}_\bullet ] \} dx .
\end{align}
The recursive formula \eqref{eq:eta} represents differential kinematics of the PCS model.
Therefore, we can obtain the relationship between $\bm{\eta}_i(s)$ and $\dot{\bm{q}}_s$ as follows: 
\begin{align}
    \label{eq:ik_final1}
     \bm{\eta}_i(s)
     =
     \bm{J}_{i}(s) \dot{\bm{q}}_s,
%     \label{eq:Jacob_pcs}
%        &\bm{J}_{i}(s)  ,
\end{align}
where $\bm{J}_{i}(s) \in \mathbb{R}^{6 \times 6n_s}$ is the Jacobian matrix at $s$ in segment $i$.

\subsubsection{Dynamics of PCS Model}
The dynamics of the continuous Cosserat rod with respect to the microsolid frame can be represented as
\begin{align}
    \label{eq:cosserat_dynamics}
    \bm{\mathcal{M}}(s)\dot{\bm{\eta}} - [\bm{\eta}{}_{\bullet}]^T \bm{\mathcal{M}}{\bm{\eta}} 
    =
    \bm{\mathcal{F}}
%    \bm{\mathcal{F}}^{\prime}_{i} + {\rm{ad}_{\bm{\xi}}^{*}} \bm{\mathcal{F}}_{i}+\bm{\mathcal{F}}_{a}+\bm{\mathcal{F}}_{e},
\end{align}
where $\bm{\mathcal{M}}(s) \in \mathbb{R}^{6 \times 6}$ is a screw inertia matrix at $s$, and
$\bm{\mathcal{F}} \in \mathbb{R}^6$ is the force applied to the microsolid.
%$\bm{\mathcal{F}}^{\prime}_{i} = \frac{\partial {\mathfrak{U}}}{\partial s}$ is a wrench of internal forces, ${\mathfrak{U}(\bm{\xi})}$ is a elastic energy of the rod, $\bm{\mathcal{F}}_{a}(s,t)$ is a distributed actuation load, and $\bm{\mathcal{F}}_{e}(s)$ is a distributed wrench of the external forces.
From this equation, the dynamics of the PCS model is derived as follows \cite{renda2018discrete}:
%In this paper, flexible tools are designed without an actuation force, $\bm{\mathcal{F}}_{a} = \bm{0}$. Therefore, \eqref{eq:cosserat_dynamics_integ2} can be rewritten as
\begin{align} 
    \label{eq:pcs_dynamics}
    \bm{\mathcal{M}}_s(\bm{q}_s) \ddot{\bm{q}}_s + \bm{C}(\bm{q}_s, \dot{\bm{q}}_s) \dot{\bm{q}}_s
    =
    \bm{\tau}_s 
\end{align}
where
\begin{align}
    \bm{\mathcal{M}}_s(\bm{q}_s)
    &=
    \int_{0}^{L_n}  \bm{J}^T \bm{\mathcal{M}}\bm{J} ds, 
    \\
    \bm{\mathcal{C}}_s(\bm{q}_s, \dot{\bm{q}}_s)
    &=
    \int_{0}^{L_n} 
    \left( 
    \bm{J}^T \bm{\mathcal{M}}\dot{\bm{J}}
    - \bm{J}^T [(\bm{J}\dot{\bm{q}})_\bullet]^T \bm{\mathcal{M}}\bm{J}
    \right)
    ds ,
    \\
    \bm{\tau}_s
    &= 
    \int_{0}^{L_n}\bm{J}^T \bm{\mathcal{F}} ds.
\end{align}
Especially in the case that only passive internal force applies inside the rod, $ \bm{\tau}_s$ is given by
\begin{align}
    \bm{\tau}_s = \bm{K}_s (\bm{q}_{s,0} - \bm{q}_s ) - \bm{D}_s \dot{\bm{q}}_s
\end{align}
where $\bm{K}_s$ and $\bm{D}_s \in \mathbb{R}^{6n_s \times 6n_s}$ are stiffness and viscosity matrices, respectively.
$\bm{q}_{s,0}$ is a neutral strain value in which no internal force is applied.

Because the structure of \eqref{eq:pcs_dynamics} is similar to that of the traditional rigid-link system, we can easily integrate the PCS model with rigid-link system as a hybrid-link system.

\begin{figure}
  \centering
  \includegraphics[width=.4\linewidth]{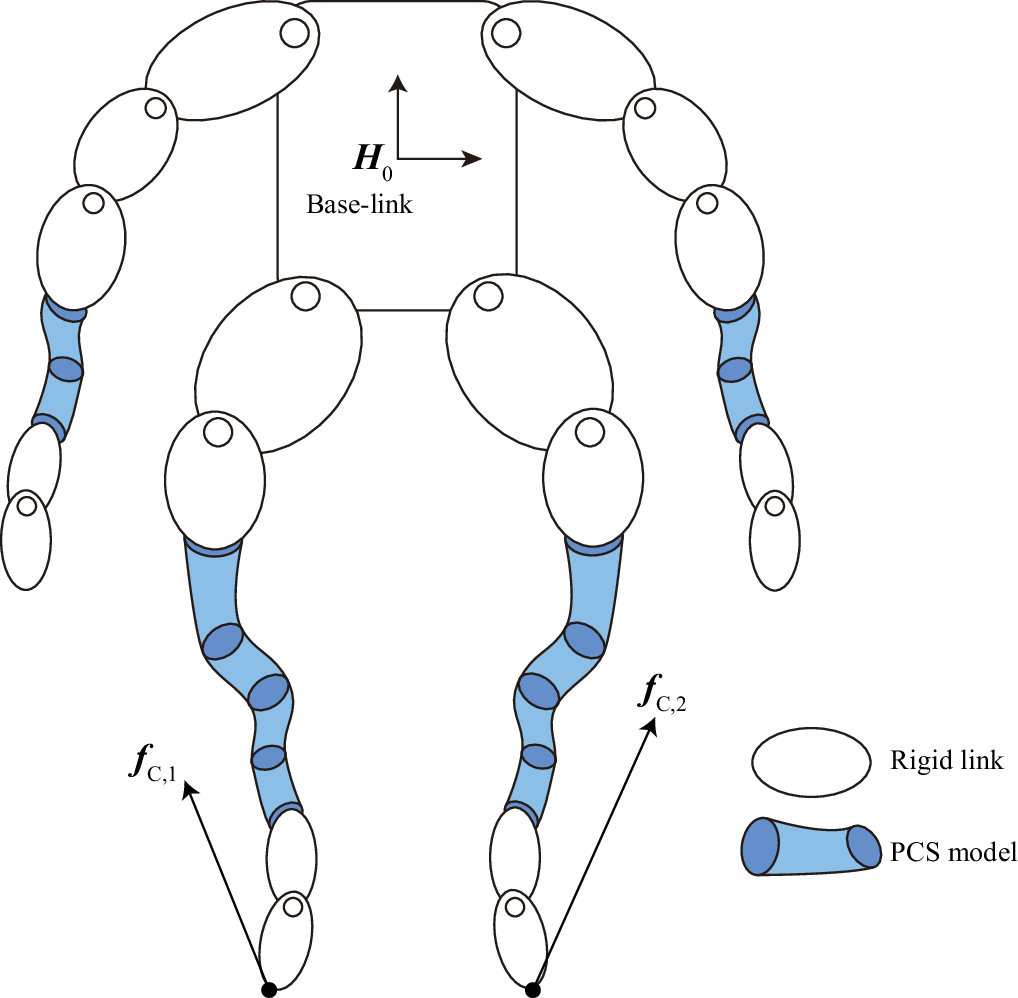}
  \caption{Conceptual illustration of a hybrid-link system that consists of rigid-links and soft-segments represented by the PCS model. We consider that the base-link is not fixed to the environment and that there are multiple contacts where $\bm{f}_{C,i}$ denote a contact force.}
  \label{fig:hybrid_link_system}
\end{figure}

\subsection{Structure of Hybrid-link System}
\figref{fig:hybrid_link_system} shows a conceptual illustration of a hybrid-link system that consists of rigid-links and soft-segments represented by the PCS model. 
As a general model, we consider that the base-link is not fixed to the environment, and that there are multiple contacts where $\bm{f}_{C,i}$ denotes a contact force.
Without loss of generality, we can define the generalized coordinate vector $\bm{q}_R$ that consists of the joint displacements of rigid-links, and $\bm{q}_S$ that consists of the strains of all soft-segments, no matter the order of these connections.
Note that $\bm{q}_R$ is an $N_R$ dimensional vector if we assume that each joint connecting rigid links is 1-DOF revolute or prismatic one, where $N_R$ is the number of joints.

Then, we define the generalized coordinate of the hybrid-link system, $\bm{q}$, and the generalized velocity vector $\bm{\psi}$ as follows:
\begin{align}
    \bm{q} &:= \left\{ \bm{H}_0, 
    \begin{bmatrix}
        \bm{q}_R^T & \bm{q}_S^T
    \end{bmatrix}^T
    \right\}, \ \mbox{and}
    \\
    \bm{\psi} &=
    \begin{bmatrix}
        \bm{\eta}_0^T & \dot{\bm{q}}_R^T & \dot{\bm{q}}_S^T 
    \end{bmatrix}^T
\end{align}
Note that we use the notation $\bm{\psi}$ for the genralized velocity, not $\dot{\bm{q}}$, because it is not directly obtained as the time-differential of $\bm{q}$.

The equation of motion of the hybrid-link system can be represented as follows:
\begin{align} 
    \label{eq:hybrid_link_dynamics} 
    & \bm{M}(\bm{q}) \dot{\bm{\psi}}+ \bm{b}(\bm{q}, \bm{\psi})
    = 
    \bm{\tau} +
    \sum_{i} \bm{J}^{T}_{C,i}\bm{f}_{C,i}.
    \\
    &\bm{M}(\bm{q})
    = 
    \begin{bmatrix}
        \bm{M}_{0} & \bm{M}_{0R} & \bm{M}_{0S} \\
        \bm{M}_{0R}^T & \bm{M}_{R} & \bm{M}_{RS} \\
        \bm{M}_{0S}^T & \bm{M}_{RS}^T & \bm{M}_{S}
    \end{bmatrix}, 
    \quad
    \bm{b}(\bm{q}, \dot{\bm{q}}) 
    =
    \begin{bmatrix}
        \bm{b}_{0} \\
        \bm{b}_{R} \\
        \bm{b}_{S}
    \end{bmatrix}, 
    \quad
    \bm{\tau}=
    \begin{bmatrix}
        \bm{0} \\
        \bm{\tau}_{R} \\
        \bm{\tau}_{S}
    \end{bmatrix},
\end{align}
where $\bm{M}(\bm{q})$ is the inertia matrix,
$\bm{b}(\bm{q}, \bm{\psi})$ is a bias vector, and 
$\bm{\tau}$ is the generalized force vector.
$\bm{M}_{\ast}$, $\bm{b}_{\ast}$ and $\bm{\tau}_{\ast}$ are a block element or vector corresponding to the base-link, rigid-link or soft-segment part.
Note that the upper part in $\bm{\tau}$ corresponding to the base-link is zero because the base-link is not fixed to the environment.
Also, $\dot{\bm{\psi}}$ is obtained from direct time-derivative of $\bm{\psi}$. For example, $\bm{M}_0$ is the inertia corresponding to $\dot{\bm{\eta}}_0$.

$\bm{J}_{C,i}$ is the Jacobian matrix of a contact point.
We can calculate the Jacobian matrix for an arbitrary point by a simple extension of the traditional algorithm in the rigid-link system, as presented in \cite{kim2022inverse}.

\subsection{Inverse Kinematics and Inverse Dynamics of Hybrid-link System}
\subsubsection{Inverse Kinematics for Motion Capture}
The outline of inverse kinematics (IK) calculation to obtain the generalized coordinates of the hybrid-link system is summarized as follows. 
Let $\bm{p}_i \ (i = 1, \cdots, M)$ denote the position vector of one of the total $M$ markers in an optical motion capture system.
We define a vector $\bm{p}$ that concatenates these position vectors as
\begin{align}
    \label{eq:marker_p_vector}
     \bm{p} = 
     \begin{bmatrix}
         {\bm{p}_1}^T & \cdots & {\bm{p}_M}^T
     \end{bmatrix}^T .
 \end{align}
In general, the IK computation is formulated as an optimization problem:
\begin{align} \label{eq:ik_problem}
     \min_{\bm{q}} \frac{1}{2} \| \widehat{\bm{p}} - \bm{p}(\bm{q}) \|_{\bm{W}_1}^2
     +
     \frac{1}{2} \| \bm{q} \|_{\bm{W}_2}^2,
\end{align}
where $\widehat{\bm{p}}$ is a measured value by the motion capture system, and $\bm{p}(\bm{q})$ is the marker position vector calculated in the hybrid-link model.
The second term is a damping factor to suppress numerical instability due to a singular configuration, which is based on the Levenberg–Marquardt method.

\subsubsection{Inverse Dynamics by Quadratic Programming}

Inverse dynamics (ID) is utilized for estimating joint torques in a musculo-skeletal model \cite{nakamura2005somatosensory,delp2007opensim} for a measured motion data.
This calculation can be easily implemented in the hybrid-link system because of the similarity of the dynamics \eqref{eq:hybrid_link_dynamics} with the traditional rigid-link system.
In the case of the floating-base system, quadratic programming (QP) is often used for simultaneously calculating the joint torques and contact forces under its constraints.

We assume that time-series data of $\bm{q}$ is obtained by the IK of the hybrid-link system, and its velocity $\bm{\psi}$ and acceleration $\dot{\bm{\psi}}$ as well.
The ID calculates $\bm{\tau}_R, \bm{\tau}_S$ and $\bm{f}_{C,i}$ from the hybrid-link dynamics \eqref{eq:hybrid_link_dynamics}.
We define vectors $\bm{x}$ and $\bm{f}_C$ that concatenate these variables as 
\begin{align}
    \bm{x} &:= 
    \begin{bmatrix}
        {\bm{\tau}_R}^T &
        {\bm{\tau}_S}^T  &
        {\bm{f}_{C}}^T
    \end{bmatrix}^T, 
    \\
    \bm{f}_C
    &:=
    \begin{bmatrix}
        {\bm{f}_{C,1}}^T & \dots & {\bm{f}_{C,n_c}}^T
    \end{bmatrix}^T
\end{align}
where $n_c$ is the number of contact points.
Given $\bm{q}, \bm{\psi}$ and $\dot{\bm{\psi}}$, we calculate the solution of $\bm{x}$ by the following QP:
\begin{align} 
    \min_{\bm{x}} 
    \frac{1}{2} \| \bm{M} \dot{\bm{\psi}} + \bm{b}
    -
    \begin{bmatrix}
        \bm{0} \\ \bm{\tau}_R \\ \bm{\tau}_S
    \end{bmatrix}
    -
    \bm{J}^{T}_{C}\bm{f}_{C} \|^2_{\bm{W}_1}  +\frac{1}{2}\| \bm{x} \|^2_{\bm{W}_2}
\end{align}
subject to
\begin{align} \nonumber
    f_{{C,i}_z} > 0,
    \quad
    \sqrt{f_{{C,i}_x}^2 + f_{{C,i}_y}^2} \leq \mu f_{{C,i}_z},
\end{align}
where $f_{C,i_x}, f_{C,i_y}$ and $f_{C,i_z}$ are the elements of $\bm{f}_{C,i}$, and $\mu$ is the maximum static friction coefficient.

\subsection{Muscle Force Estimation}
\subsubsection{Muscle Force Optimization}
From the joint torques estimated by the ID, we can estimate muscle forces that actuate each joint.
Because a human joint is redundantly actuated by multiple muscles, muscle forces $\bm{f}$ are usually estimated by an optimization given as follows \cite{nakamura2005somatosensory}:
\begin{align}
    \label{eq:ms_optimization}
    \min_{\bm{f}} \frac{1}{2} \| \bm{\tau}_R - \bm{J}_{\ell}^T \bm{f} \|^2_{\bm{W}_1}
    +
    \frac{1}{2} \| \bm{f} - \bm{f}_{ref} \|^{2}_{\bm{W}_2}
\end{align}
subject to
\begin{align}
    \bm{0} \le \bm{f} \le \bm{f}_{\max}
\end{align}
where $\bm{J}_{\ell}$ is the Jacobian matrix of the muscle length, and $\bm{f}_{\max}$ is the maximum muscle force calculated from Hill's model \cite{hill1938heat}.
$\bm{f}_{ref}$ is a reference value calculated from Hill's model as well if muscle activity can be obtained from electromyography (EMG) sensors.

\begin{figure}[tbp]
  \centering
  \includegraphics[width=0.5\linewidth]{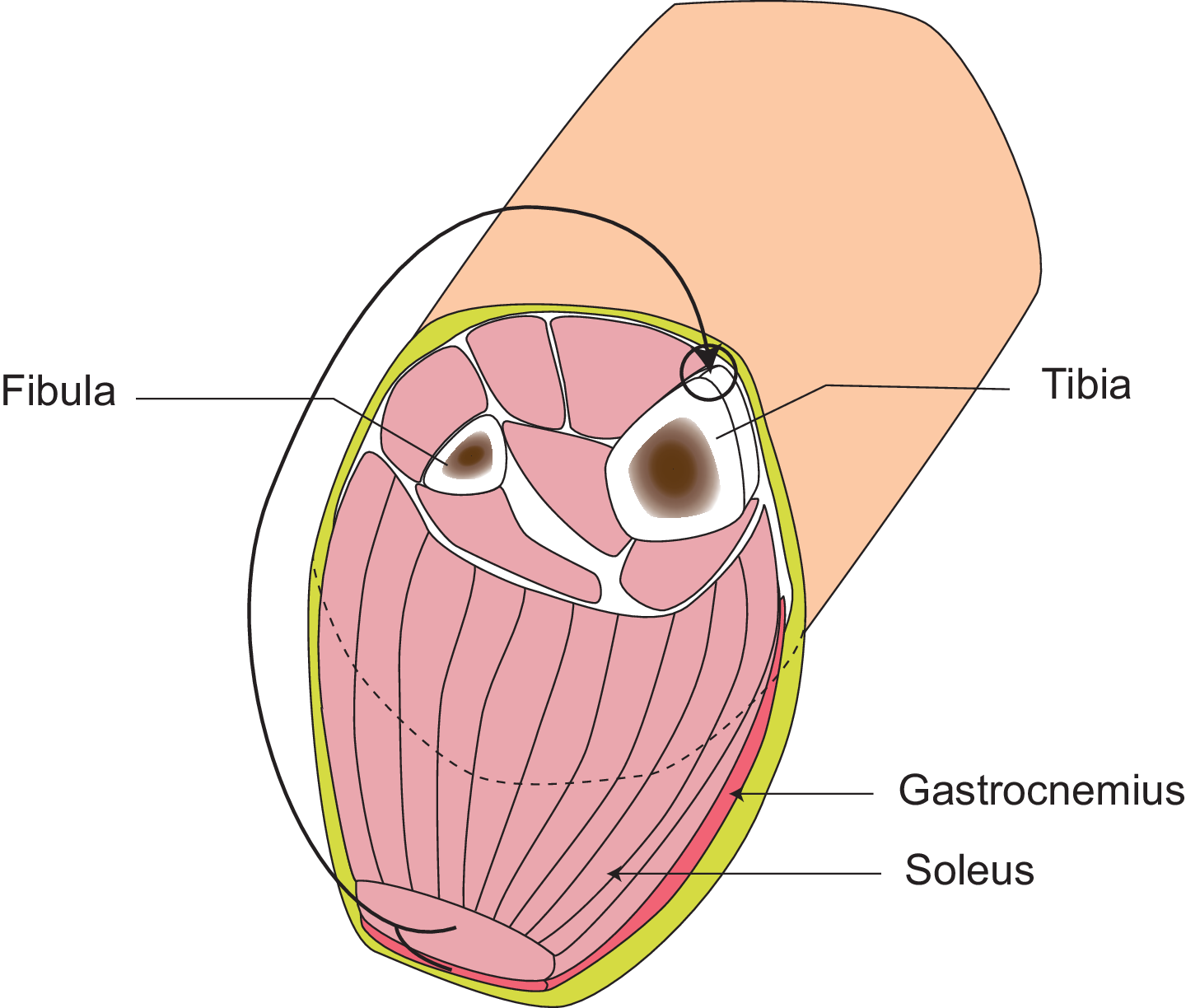}
  \caption{\textcolor{black}{Schemetics for the transtibial amputation surgery by myodesis focused on muscles}}
  \label{fig:amputation_myodesis_method}
\end{figure}

\subsubsection{Muscle Amputation Modeling}
In this study, we use a musculo-skeletal model developed by \cite{nakamura2005somatosensory} and modified it to consider a muscle amputation.
It is known that there are two methods for amputation surgery: myodesis and myoplasty \cite{geertzen2019myodesis}. 
Myodesis is a method to drill a hole in a bone and secure muscles inside the hole whereas myoplasty stitches severed muscles over the end of the bone. 
Each method has advantages and limits, and a medical doctor chooses a method depending on various factors. 
Therefore, it is hard to unify the muscle model of amputation stump and need to design based on individual characteristic for accurate modeling. 
% Therefore, it is hard to normalize the arrangement of the muscles on the amputation stump.

\begin{figure}[tbp]
  \centering
  \includegraphics[width=.4\linewidth]{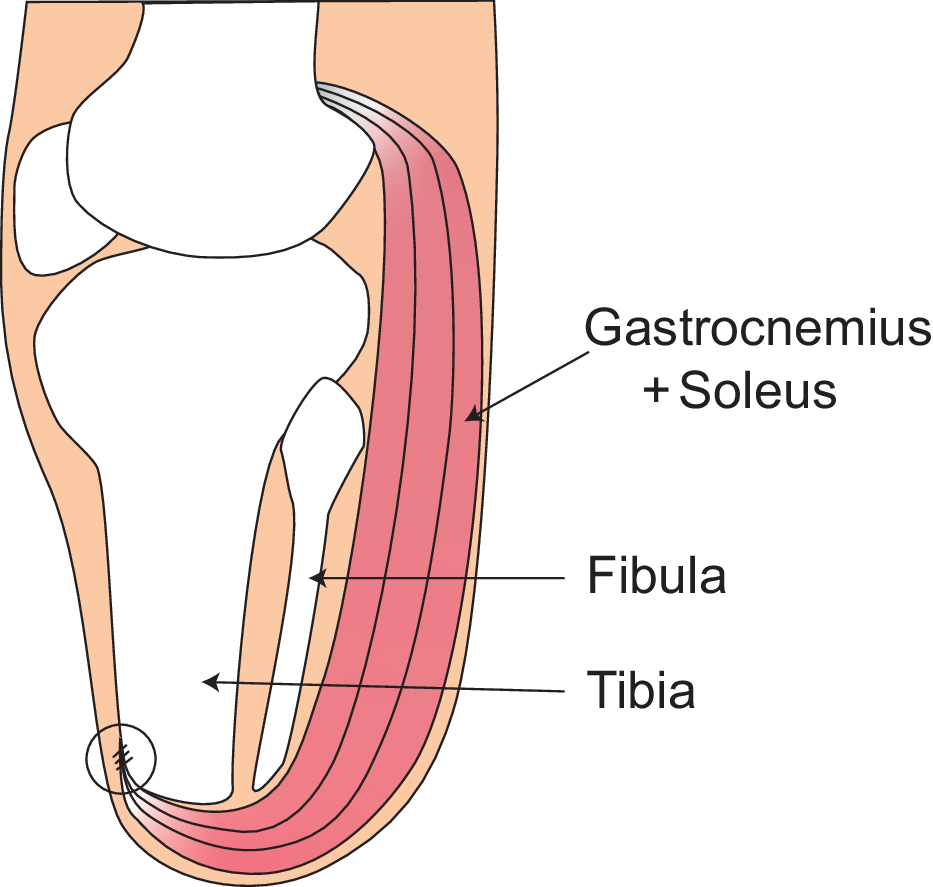}
  \caption{\textcolor{black}{Schematic of the side view of the MRI scanning for leg with transtibial amputation operated by myodesis.}}
  \label{fig:amputation_myodesis_mri}
\end{figure}
In this study, we found that our subject had a myodesis surgery from MRI scanning of the subject and designed the amputation stump based on the observation.
\figref{fig:amputation_myodesis_method} illustrates the schematics for myodesis, focused on the muscle amputations, drawn based on \cite{brown2014outcomes}. 
The muscle amputation is summarized as follows:
\begin{itemize}
    \item Anterior compartment musculature was removed.
    \item Deep posterior musculature (Gastrocnemius and Soleus) was transected and secured to the tibia via small holes.
\end{itemize}
If the transtibial amputation is operated by myodesis, the side view of the MRI scanning of the amputated leg can be illustrated as \figref{fig:amputation_myodesis_mri}. 
Due to the personal information protection, we cannot include the actual MRI image in this paper. 
Instead, we observed that the gastrocnemius and the soleus shown in the MRI image had an arrangement similar to \figref{fig:amputation_myodesis_mri}. 

\begin{figure}[tbp]
  \centering
  \includegraphics[width=0.4\linewidth]{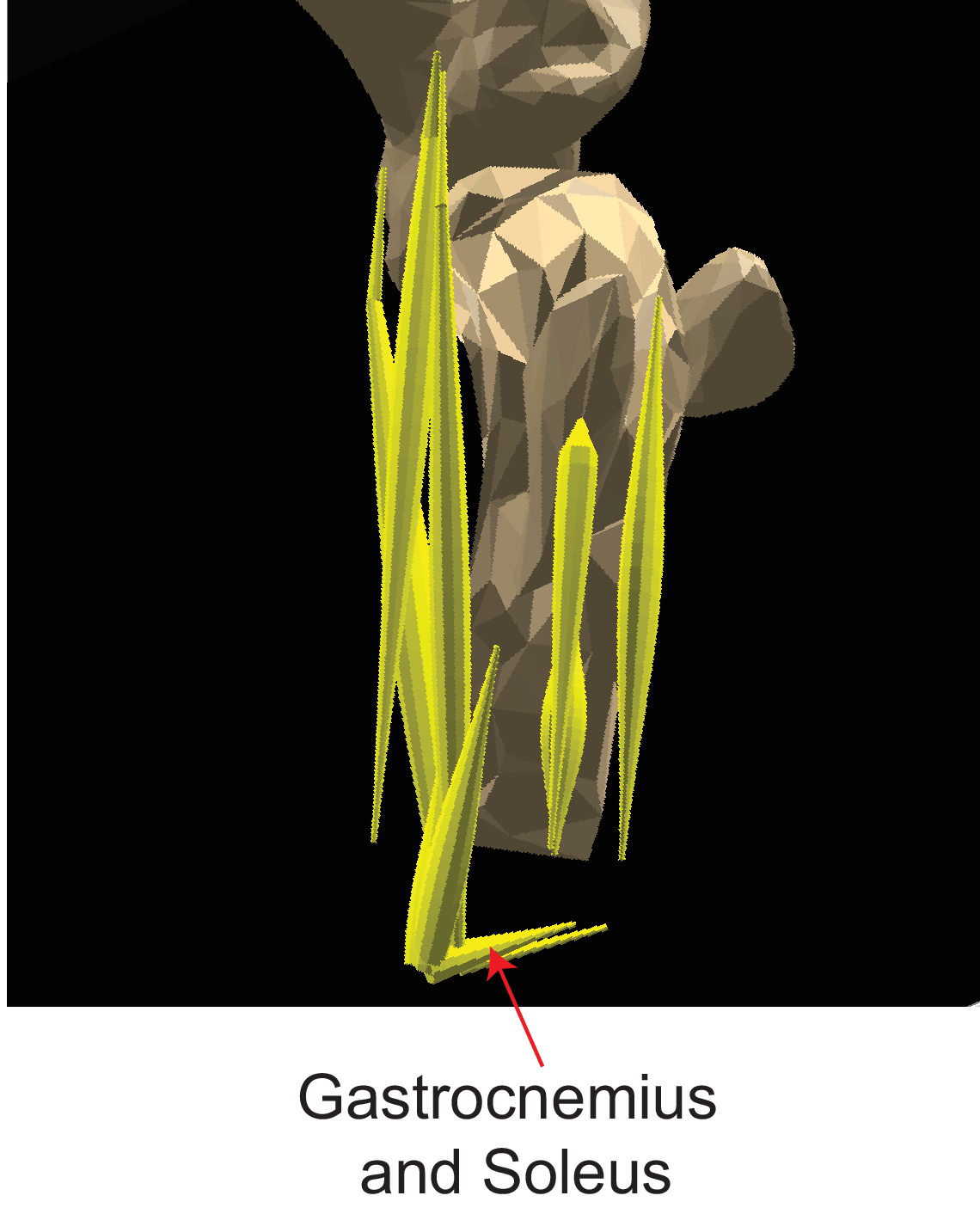}
  \caption{Arrangement of the transected muscles based on the MRI image of the subject.}
  \label{fig:muscle_before_after}
\end{figure}

According to the aforementioned observation, we design the transected muscles as follows:
\begin{itemize}
    \item Anterior compartment muscules (extensor digitorum longus, peroneus longus, plantaris, and tibialis anterior) are transected on the amputation stump and does not secured in any bone.
    \item Deep posterior musculature (Gastrocnemius and Soleus) was transected longer than the amputation stump. Also, the end of the muscles are secured to the front of the tibia. Therefore, these muscles has the shape as wrapping the tibia stump.
\end{itemize}
\figref{fig:muscle_before_after} shows the the muscle arrangement of an MRI-based designed amputation stump, in which each muscle is visualized as a simple thin wire without volume to show its arrangement. 
%\figref{fig:muscle_before_after} (a) shows the muscle arrangement with simple connection. 
%In this design of amputation stump, the end of the transected muscles are converged at a single point under the bone stump. 
%Instead, \figref{fig:muscle_before_after} (b) shows the muscle arrangement of a MRI-based designed amputation stump. 
In this figure, the gastrocnemius and soleus muscles are wrapping tibia stump, as their endpoints are secured in the front of the tibia. 

%% file: contents/results.tex
\begin{figure}
    \centering
    \includegraphics[width=.6\linewidth]{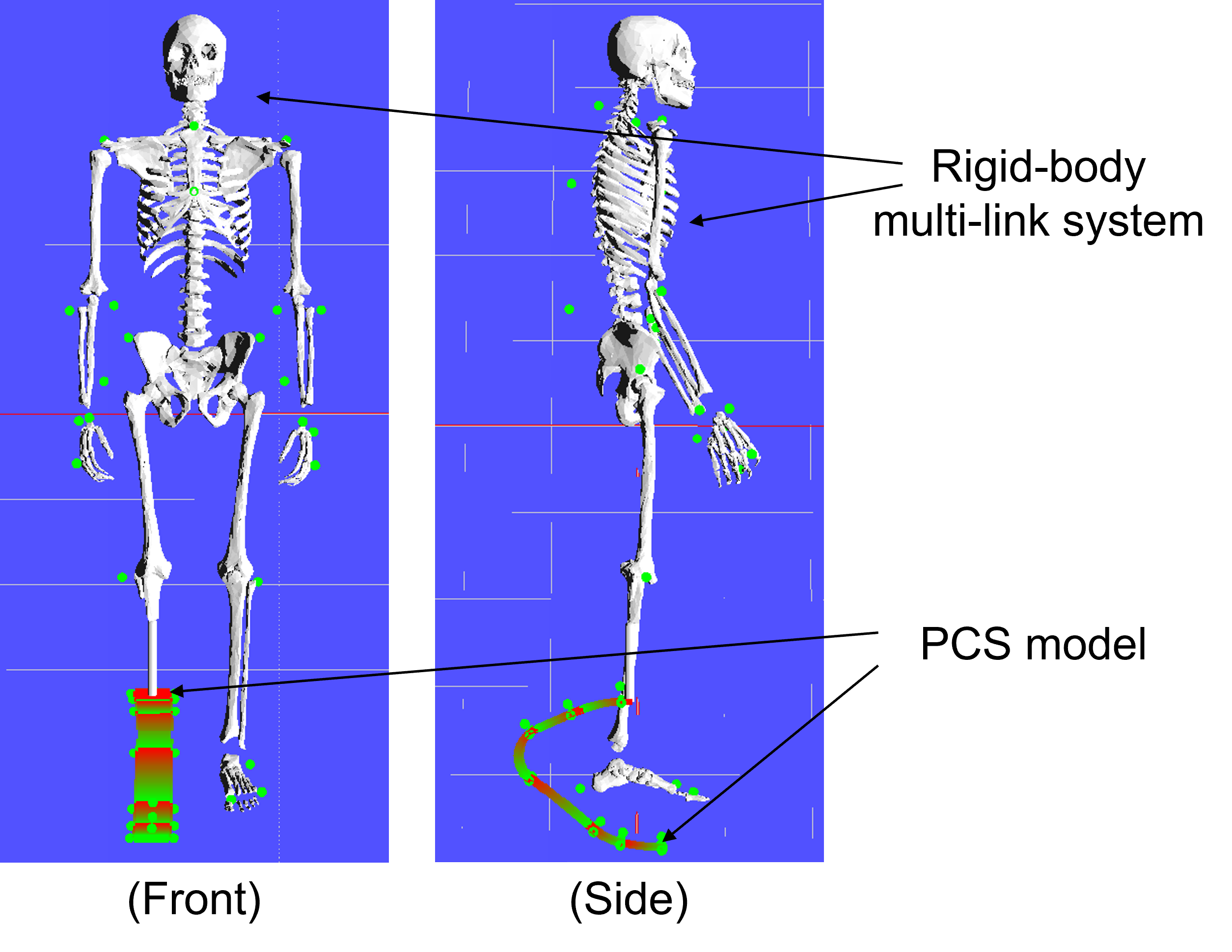}
    \caption{Hybrid-link model that integrates human skeleton and a leaf-spring prosthesis leg.}
    \label{fig:hybrid_skeletal_model}
\end{figure}

\begin{figure}
    \centering
    \includegraphics[width=.45\linewidth]{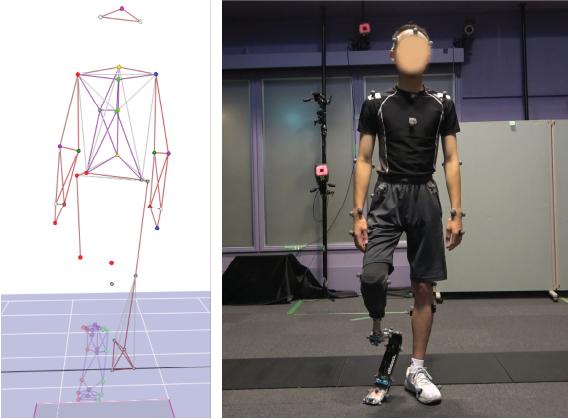}
    \includegraphics[width=.45\linewidth]{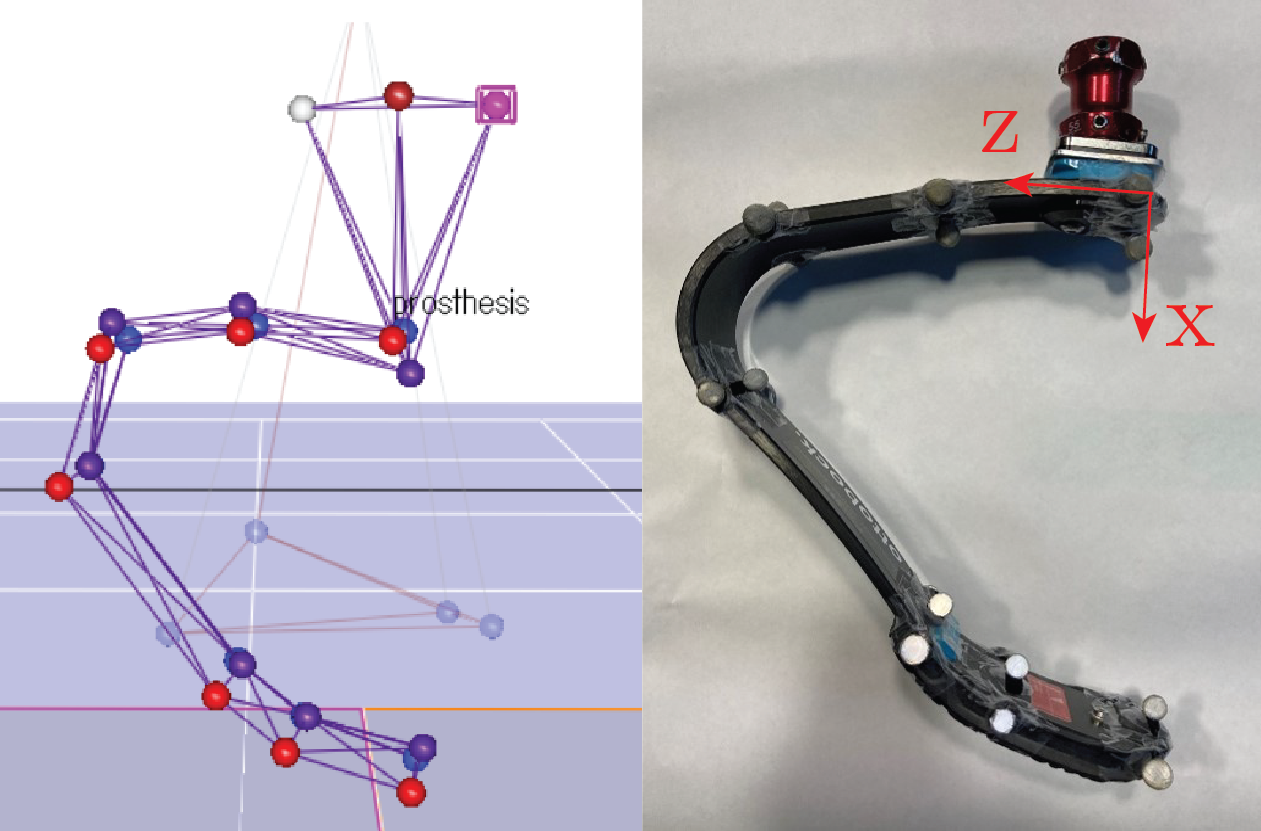}
    \caption{Layout of optical markers for motion capture measurement.}
    \label{fig:marker_position_reduced}
\end{figure}

\section{Experiments and Results\label{sect:results}}
\subsection{Motion Capture Measurement and Inverse Kinematics \label
{sect:hybrid_ik}} 
We built a hybrid-link model for an athlete with a leaf-spring prosthesis leg, as shown in \figref{fig:hybrid_skeletal_model}.
The rigid part for the human skeletal model without right lower-limb is based on the musculoskeletal model developed in \cite{nakamura2005somatosensory}.
Different from \cite{nakamura2005somatosensory}, our model consists 50 rigid links and proper joints with 130 degrees of freedom (DOFs). 
A socket used to relate a severed human leg and prosthesis was designed as a rigid-link without DOF.
The prosthetic leg (1E91 Runner by Ottobock) was designed as 6-segment PCS model. 
The number of segments was determined based on the previous report \cite{shimane2022application}, where the change ratio of strain along the central axis coordinate was large. 
We assume that each segment has only angular strain with 3 DOFs ignoring linear strain while $\bm{q}_{s,0}$ is calculated so that each segment has 6 DOFs for a precise modeling.

%However, 3 markers on head were not used for IK since the size of head is largely different from people to people so the scaling is difficult.

As a preliminary experiment \cite{kim2022inverse}, we obtained a walking motion of a subject wearing prosthesis. 
In the motion capture measurement, 51 optical markers were attached: 30 of them on human links and the remaining 21 on the prosthesis. 
\figref{fig:marker_position_reduced} shows the layout of the markers.
The subject was a healthy 15 year-old male who had undergone the right lower-extremity amputation approximately 13 cm from the knee. 
This subject wore a prosthesis on his right-side and walked on a treadmill with constant velocities 2.4km/h. 
The motion was captured with 200 Hz frequency.
%The subject was provided informed consent to the protocol approved by the institutional review board before the experiment. 
%Ethics approval for the experimental procedures was granted by the Human Research Committee at Graduation School of Information Science and Technology, the University of Tokyo (approval number: UT-IST-RE-220209). 

\begin{figure}
    \centering
    \includegraphics[width=1.0\linewidth]{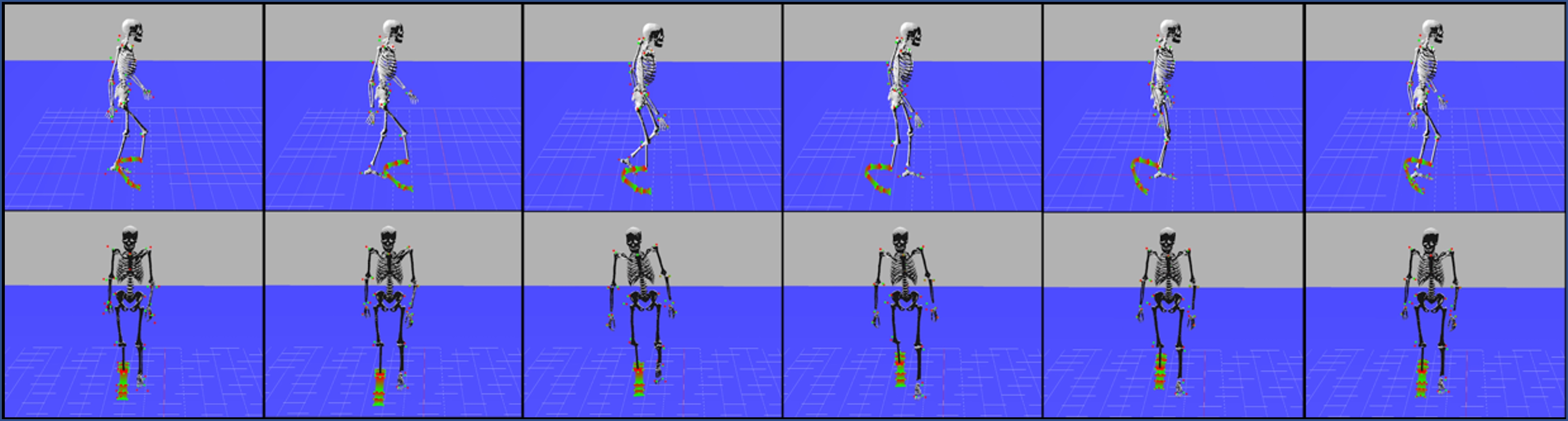}
    \caption{Walking motion reconstructed by the inverse kinematics of the hybrid-link system.}
    \label{fig:ik_result_hybrid_reduced}
\end{figure}

\figref{fig:ik_result_hybrid_reduced} shows a walking motion reconstructed by the IK of the hybrid-link system.
It is observed that the subject's motion is appropriately calculated by the IK.

\begin{figure}
    \centering
    \includegraphics[width=.5\linewidth]{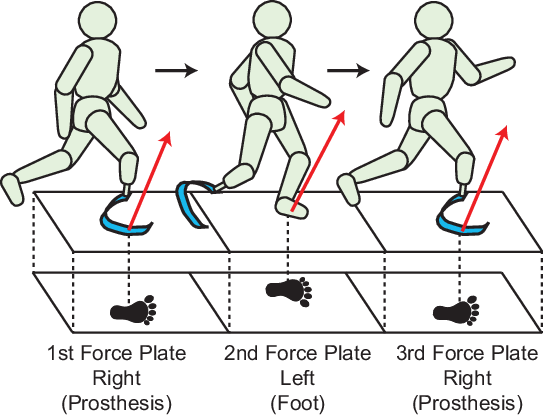}
    \caption{A subject was asked to walk or run with a single stride on each force plate in the experiment}
    \label{fig:exp_setup}
\end{figure}

\begin{figure}
    \centering
    \includegraphics[width=.5\linewidth]{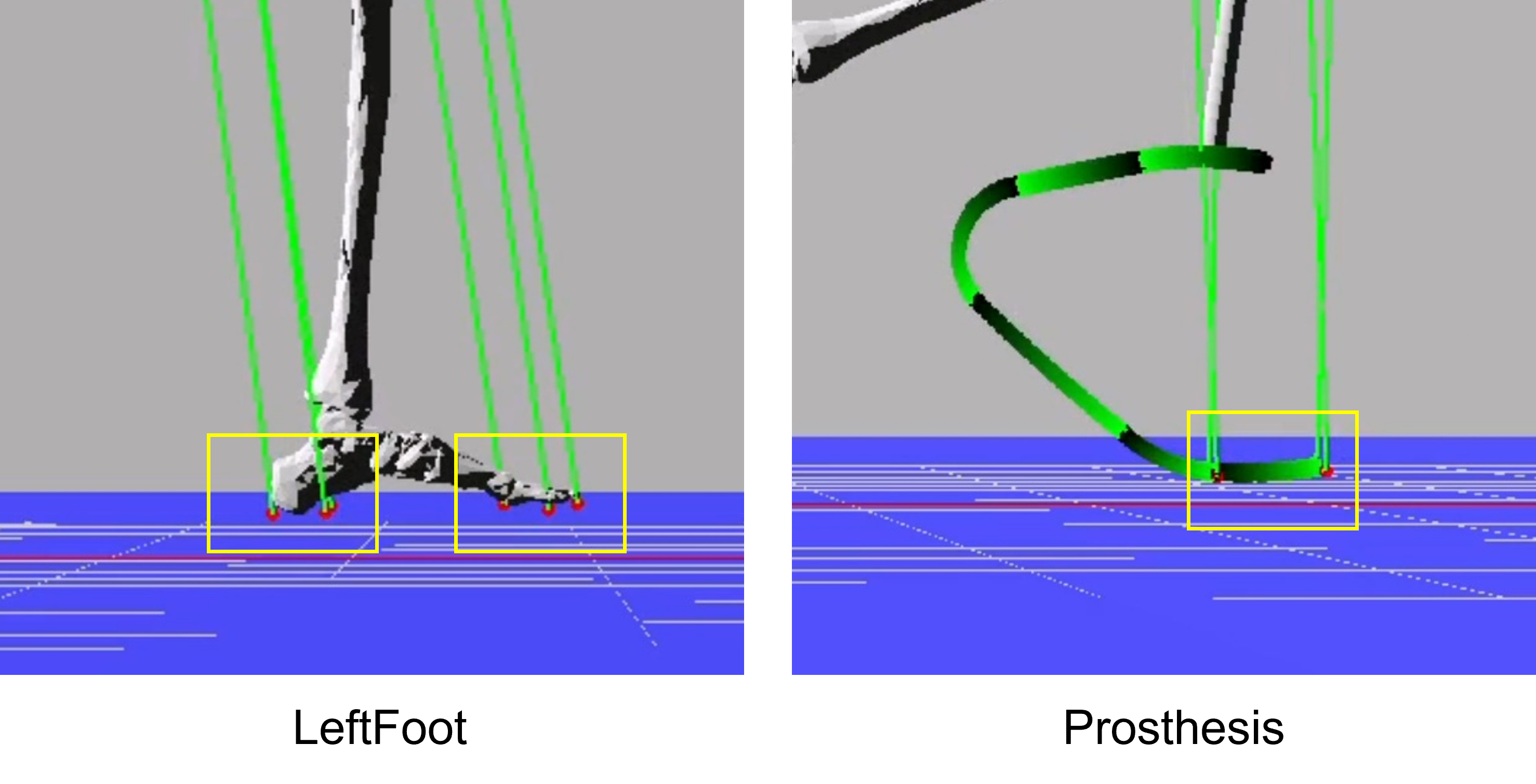}
    \caption{Setting of contact points in the inverse dynamics.}
    \label{fig:contact_point_setting}
\end{figure}

\subsection{Validation of Ground Reaction Force Estimation}
\begin{figure}
    \centering
    \includegraphics[width=0.7\linewidth]{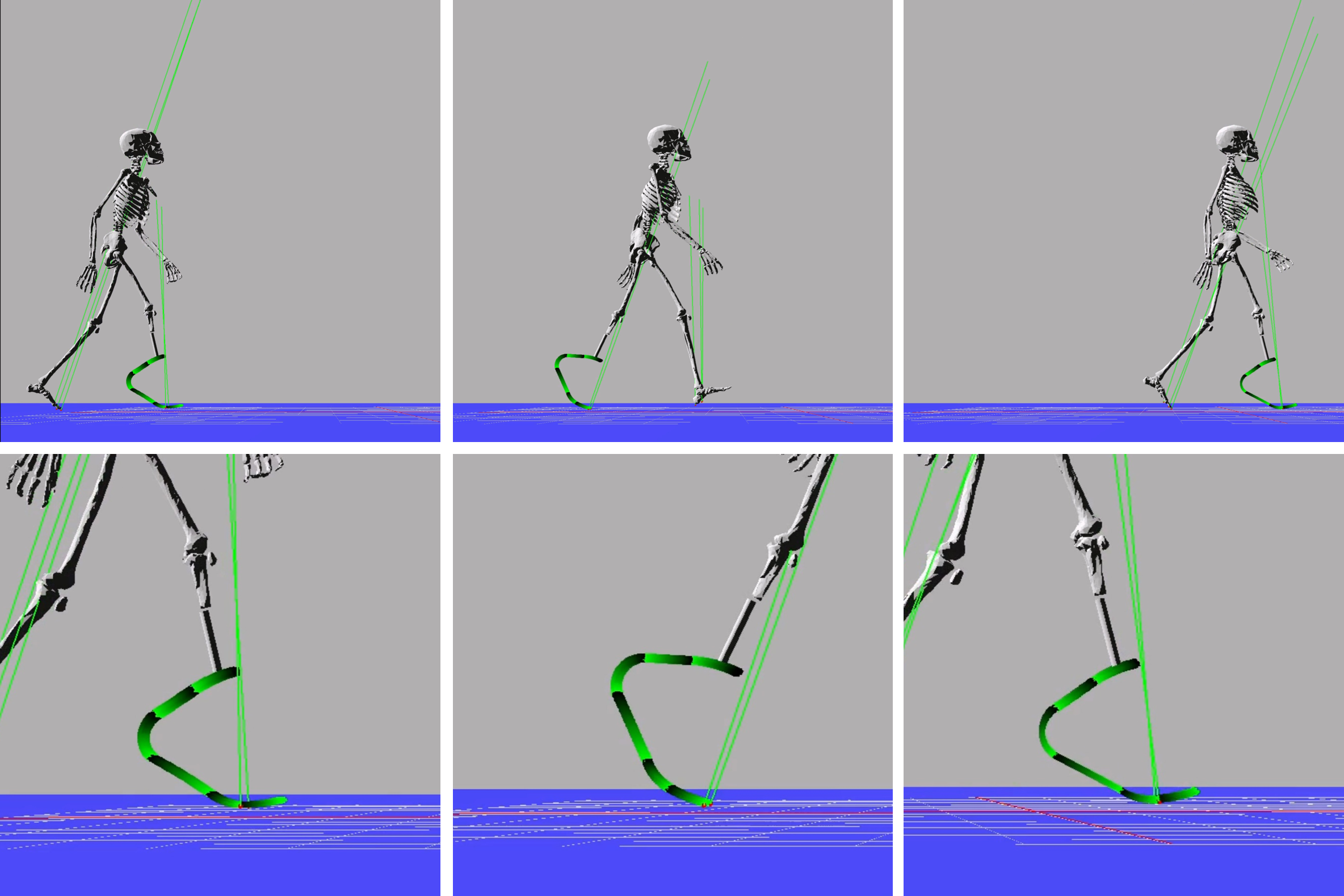}
    \caption{Reconstructed walking motion and estimated ground reaction force.}
    \label{fig:result_id_walk}
\end{figure}

\begin{figure}
    \centering
    \includegraphics[width=0.7\linewidth]{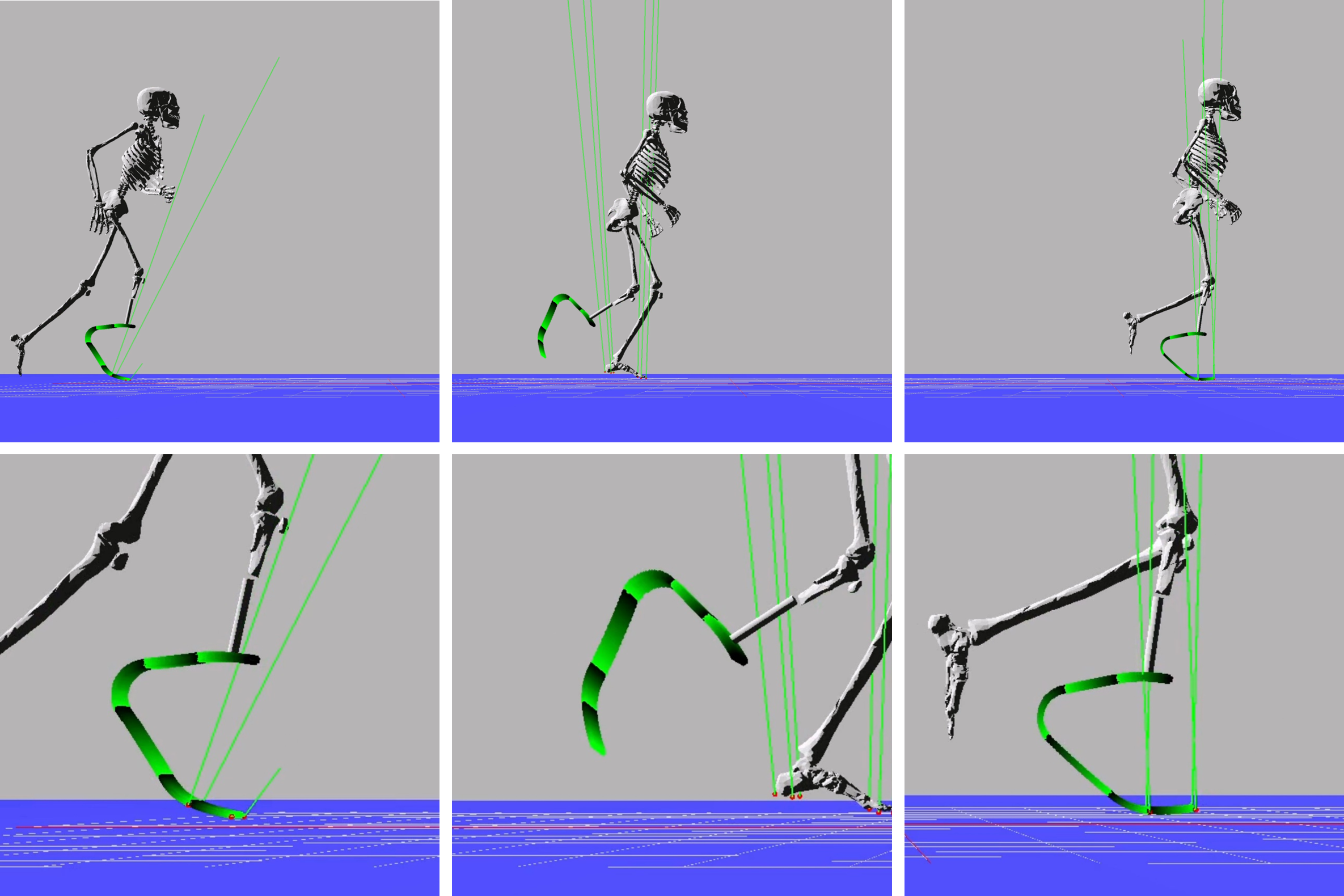}
    \caption{Reconstructed running motion and estimated ground reaction force.}
    \label{fig:result_id_run}
\end{figure}

We validate the ID calculation of the hybrid-link system using walking and running motions obtained from another motion capture measurement.  
%A healthy 17-year-old male who have undergone the right lower-extremity amputation approximately 13 cm from the knee, and wore a prosthesis on his right side, volunteered. 
%During the experiment, 33 optical markers were attached to the human joints and 21 markers were attached to the running-specific prosthesis. 
%
While the experiment reported in Section \ref{sect:hybrid_ik} was on a treadmill, the subject was walking or running on three force plates in this experiment to measure the ground reaction force (GRF). 
As shown in \figref{fig:exp_setup}, the subject was asked to walk or run with a single stride on each force plate.
%As in our previous report \cite{kim2022inverse}, the scaling of the marker position for the human part was applied because of difficulties in measuring the direct marker position. 
%Furthermore, the ground reaction force of the participant during motion was measured using three force plates placed under the steps of the participant. 
%he participant was guided to walk or run on each force plate with a single step while shifting his leg, as illustrated in Fig. \ref{fig:exp_setup} (b).
%alking and running motion were performed with this guidance at the participant's own pace.
In the ID, we set candidates of the contact points: 6 points for the left foot and 4 points for the prosthesis, as shown in Fig. \ref{fig:contact_point_setting}.
Figs. \ref{fig:result_id_walk} and \ref{fig:result_id_run} show reconstructed walking and running motions obtained from the experiment, respectively.
In each figure, the green lines indicate the contact forces estimated by the ID calculation.
% For gait motion of the athlete, the ground reaction force is the only contact force affect to the hybrid system. 
% This means, only ground reaction force is need to be measured to evaluate inverse dynamics result with contact force.

\begin{figure}[!ht]
    \centering
    \includegraphics[width=\linewidth]{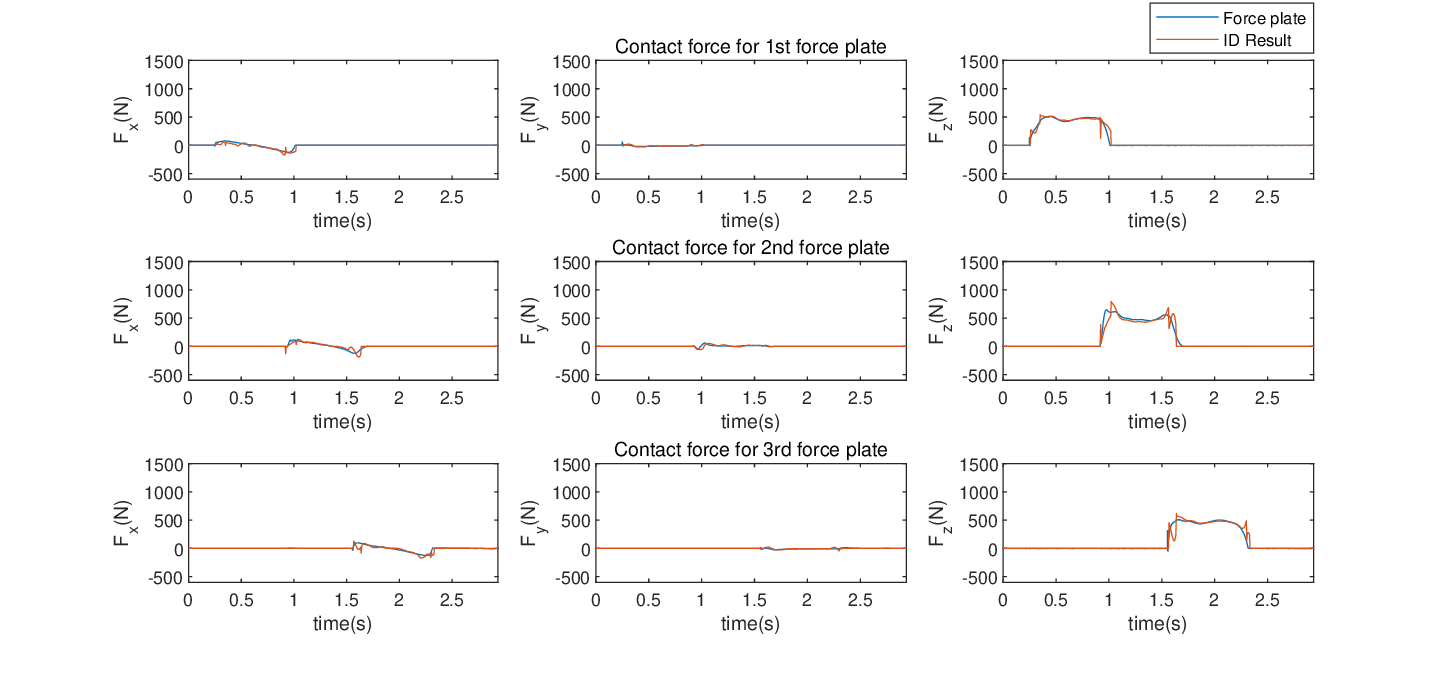}
    \caption{Estimated contact force and its ground truth along xyz axes in the walking motion.}
    \label{fig:result_id_plot_walk}
\end{figure}

\begin{figure}[!ht]
    \centering
    \includegraphics[width=\linewidth]{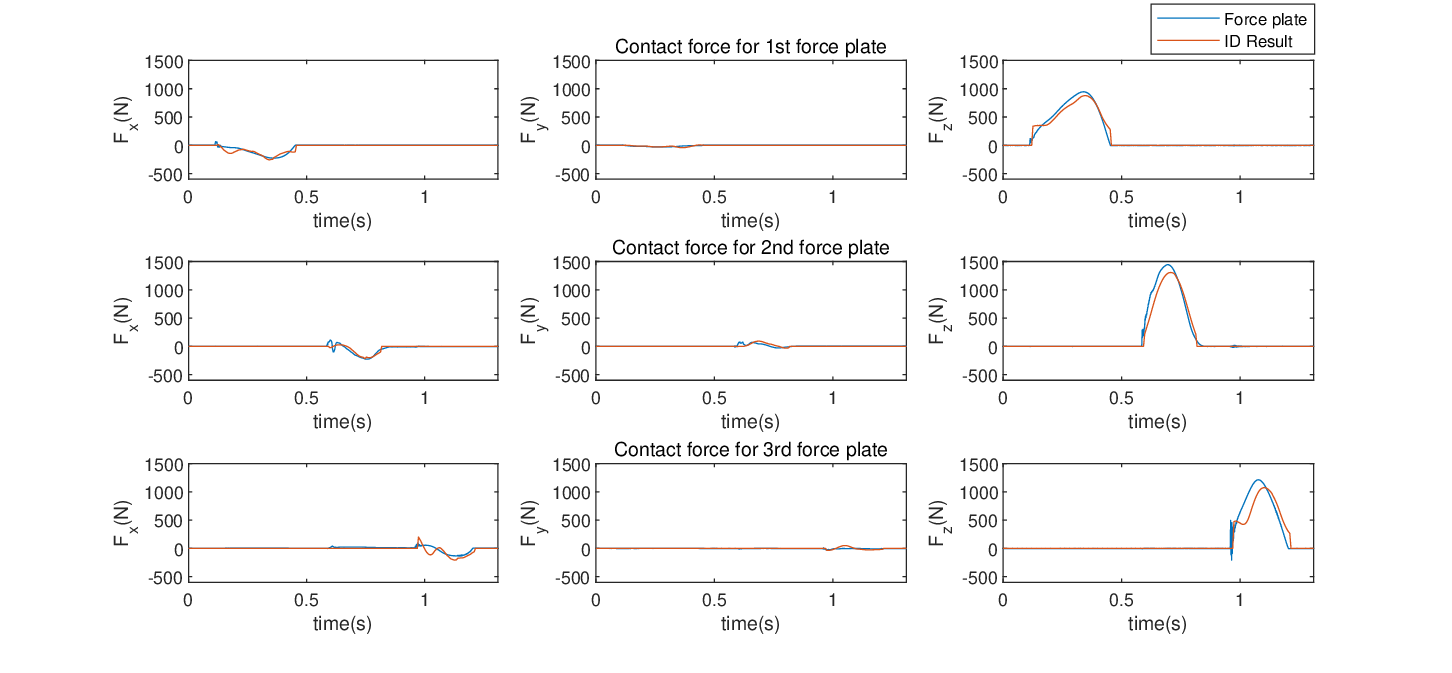}
    \caption{Estimated contact force and its ground truth along xyz axes in the running motion.}
    \label{fig:result_id_plot_run}
\end{figure}

% \begin{figure*}[!ht]
%     \centering
%     \includegraphics[width=.8\linewidth]{figure/id_result_run.eps}
%     \caption{Estimated contact force for each force plate along xyz axis for running motion}
%     \label{fig:result_id_run}
% \end{figure*}

The ID estimates the joint torques of the human skeleton and generalized force of the prosthesis represented by the PCS model, in addition to the GRF.
However, it is difficult to directly measure the first two values.
Therefore, we validate the ID by comparing the estimated GRF with the ground truth obtained from the force plates.
Figs. \ref{fig:result_id_plot_walk} and \ref{fig:result_id_plot_run} show the results of the GRF estimation in the walking and running motions, respectively.
In each figure, the blue and red lines indicate the ground truth and estimated values, respectively.
Overall, it is observed that the estimated results are sufficiently close to the ground truth.

\begin{figure}[tbp]
    \centering
    \includegraphics[width=0.36\linewidth]{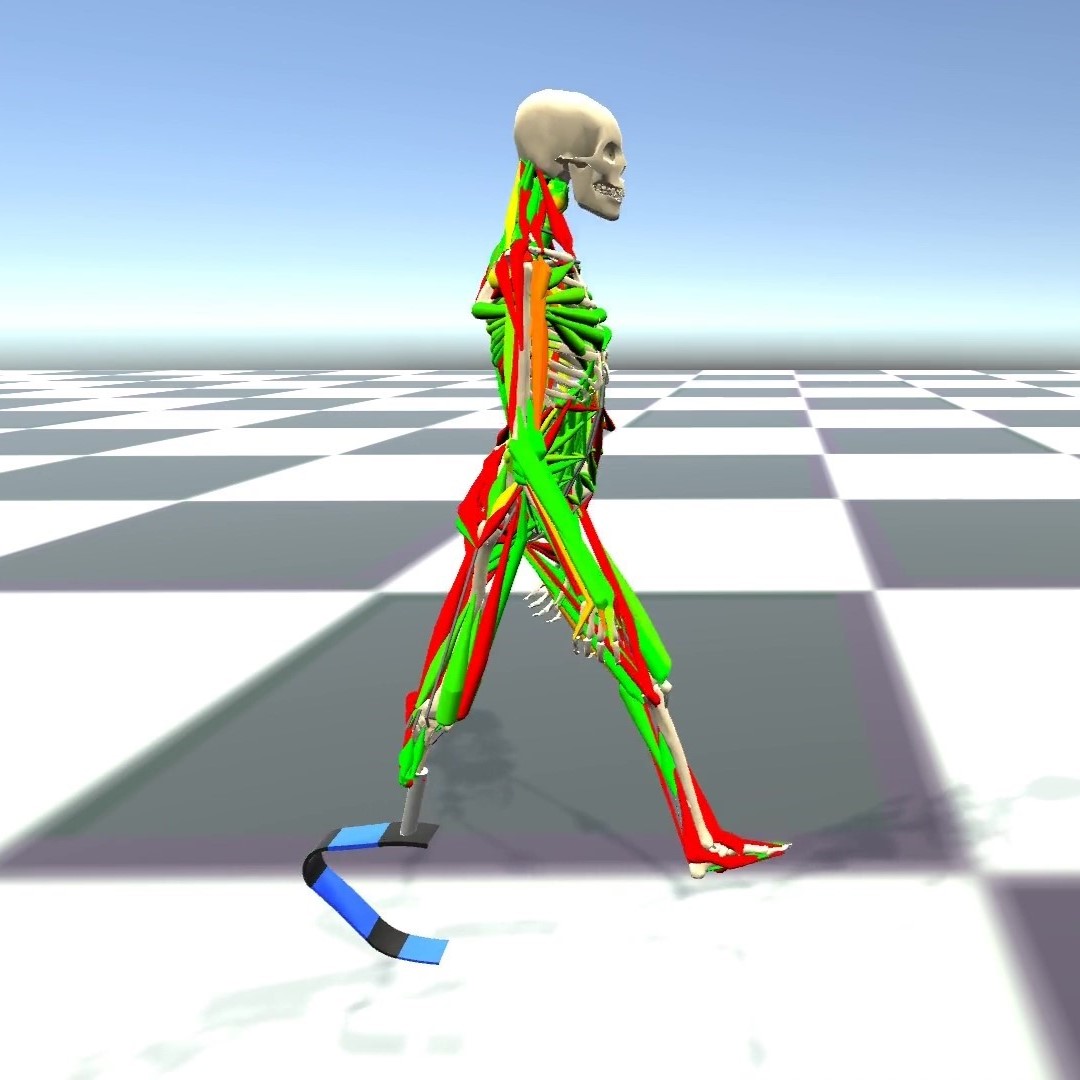}
    \includegraphics[width=0.5\linewidth]{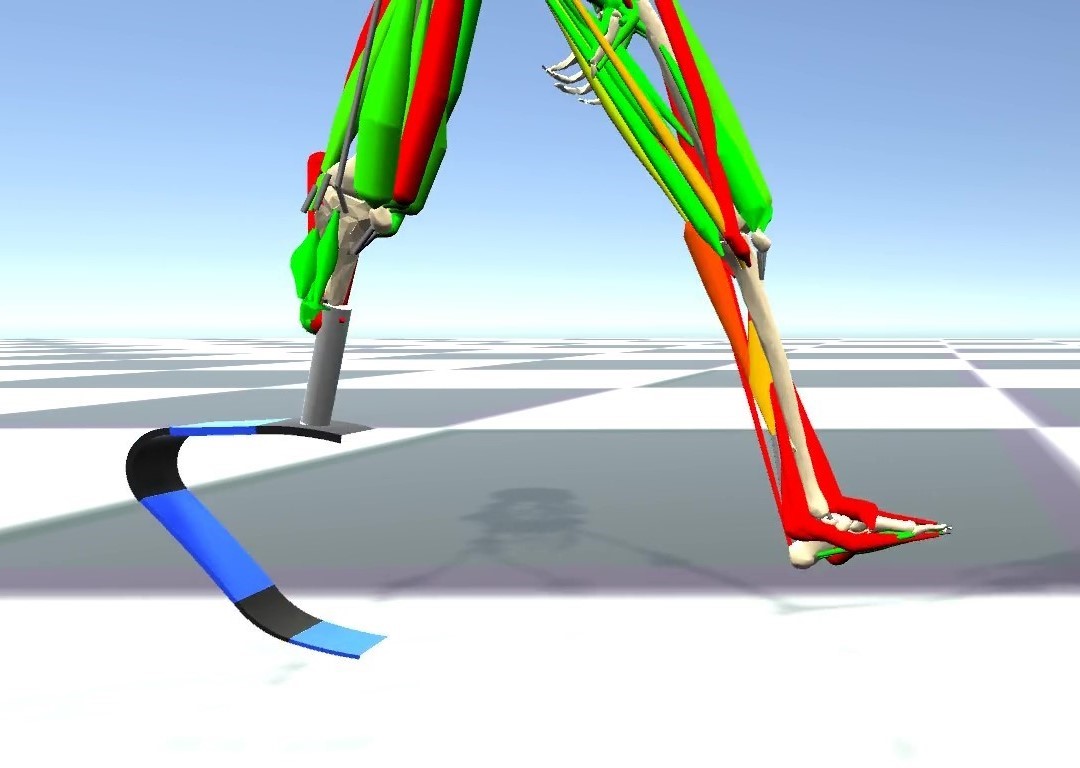}
    \caption{Visualization of estimated muscle activities, in which the red part indicates higher activity, and the green part indicates lower activity.}
    \label{fig:hybrid_msmodel_visualized}
\end{figure}
\subsection{Comparison between Muscle Force Estimation by Optimization and that based on EMG Signal}
\figref{fig:hybrid_msmodel_visualized} and accompanying video show a visualization of the estimated muscle activity, in which the red part indicates higher activity, and the green part indicates lower activity.
For a quantitative verification, we compared the estimated muscle force by the optimization and the estimated value using the EMG signal (see Appendix \ref{apdx:msforce_calc}).
For the comparison, we did not use the EMG signal in the optimization \eqref{eq:ms_optimization}, setting $\bm{W}_2 = \bm{O}$.
This would degrade the precision of the estimation; however, we excluded the EMG signal factor from the estimation for a pure comparison as a trial.

\begin{figure}[tbp]
    \centering
    \includegraphics[width=0.6\linewidth]{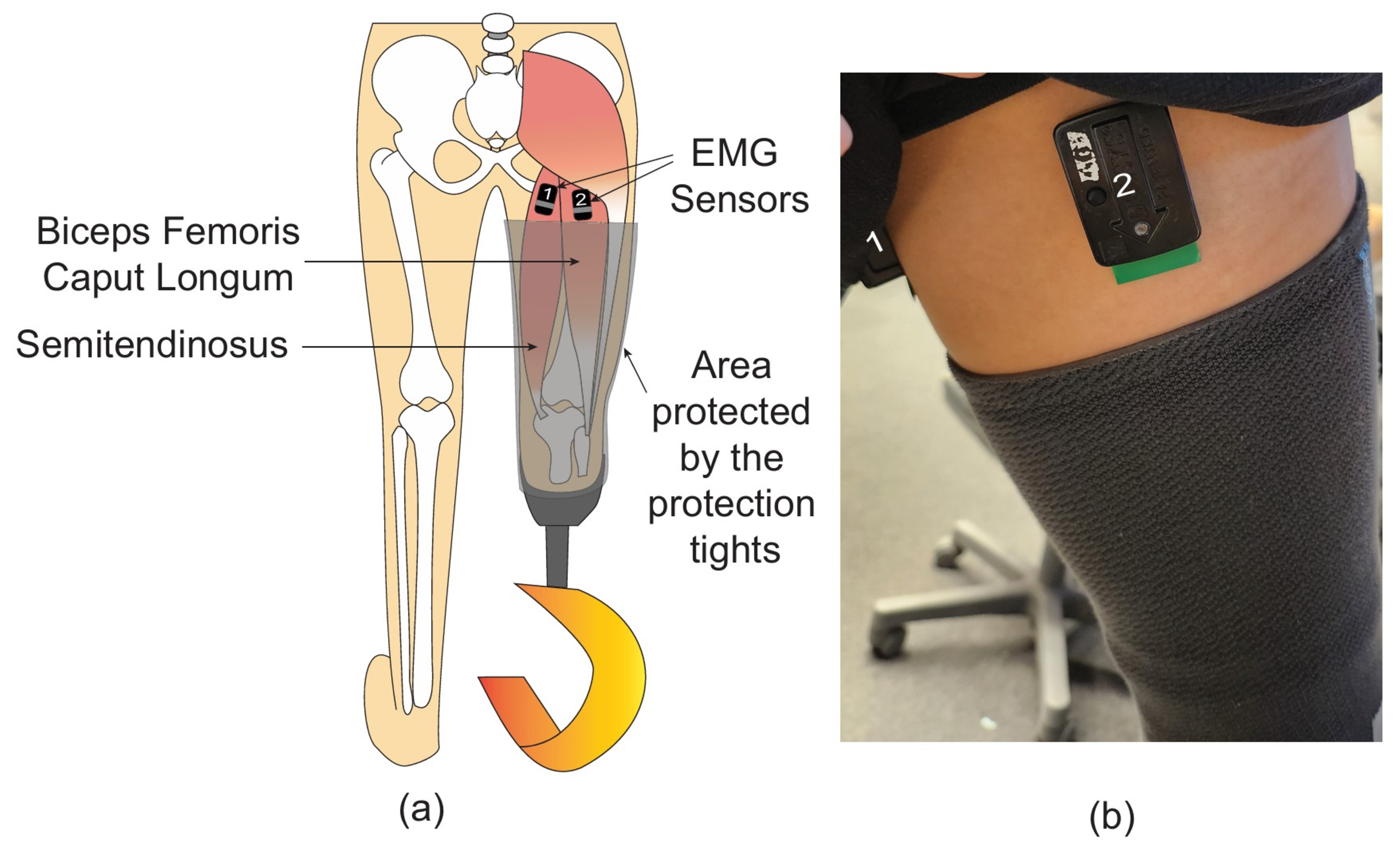}
    \caption{Location of the muscle and EMG sensors for the experiment}
    \label{fig:emg_location}
\end{figure}
During the experiment, the subject wore tights to protect the connecting part between the socket and amputated leg. 
If the air enters inside the tights, the subject could not perform a running motion. 
Therefore, we need to attach EMG sensors without touching protection tights. 
From this reason, we attached two EMG sensors on the posterior leg, as shown in \figref{fig:emg_location}, which can be used to obtain a reliable result as possible. 
In \figref{fig:emg_location}, EMG sensor 1 is attached on the Semitendinosus, and EMG sensor 2 is attached on the bicepts femoris caput longum muscle.

\if0
For experiment to obtain EMG signal, we attached the EMG sensors on the muscles of the participant and measured the running motion of the participant by motion capture system with force plate. Running motion is performed with 3 steps on the force plate same as \figref{fig:experiment_id}.
The location of EMG sensors used in this subsection is illustrated in \figref{fig:emg_location}. 
We used two EMG sensors shown in \figref{fig:emg_location} for the validation although more EMG sensors are attached in the measurement.
\fi

Ideally, we should compare the results of transected muscles to validate the model of amputation. However, it was hard to attach EMG sensors on the amputated muscles inside the protection tights due to the safety problem.
Instead, we compared the results of bicepts femoris caput longum muscle and semitendinosus muscle.
Although these muscles are not transected ones, they also generate the knee joint torque, which is the most affected joint from the transtibial muscle amputation. 
If the change of transected muscle tensions affects the knee joint torque, the action of these muscles will be altered together. 
Therefore, we consider that the evaluation of these two muscles is still reasonable.

\begin{figure}[!ht]
    \centering
    \includegraphics[width=0.6\linewidth]{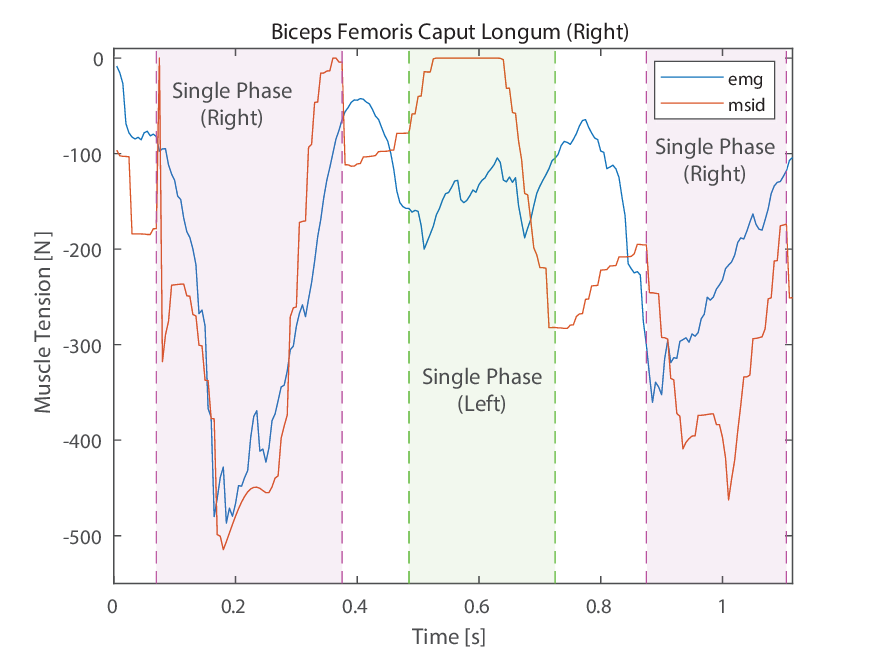}
    \caption{Muscle tension comparison between the estimations by the optimization and the EMG signal: Biceps Femoris Caput Longum}
    \label{fig:bicepsfemoriscaputlongum}
\end{figure}

\begin{figure}[!ht]
    \centering
    \includegraphics[width=0.6\linewidth]{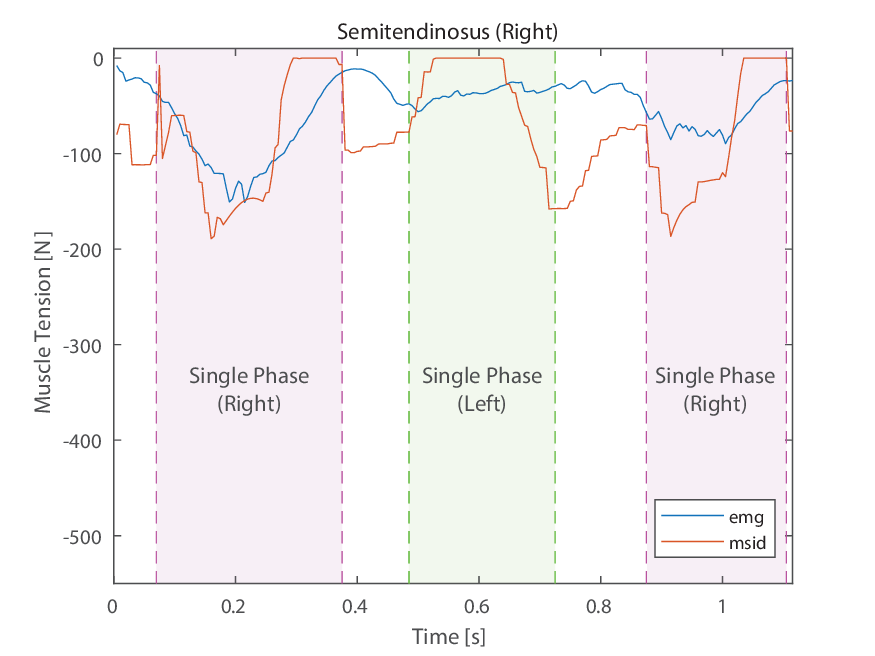}
    \caption{Muscle tension comparison between the estimations by the optimization and the EMG signal: Semitendinosus}
    \label{fig:semitendinosus}
\end{figure}

Figs. \ref{fig:bicepsfemoriscaputlongum} and \ref{fig:semitendinosus} show the estimated muscle tension forces by optimization (indicated by the red line) and those by EMG signal (indicated by the blue line).
In each figure, pink areas indicate the single phase with the right leg, in which the subject stood with the prosthesis on the force plate whereas green areas indicate the single phase with the left leg.
Note that it is well-known that these two estimation does not have a precise agreement.
Usually, it can be validated if a similar profile is obtained.
In Figs. \ref{fig:bicepsfemoriscaputlongum} and \ref{fig:semitendinosus}, we can observe a similarity between the two results in the first phase.
In the second phase, a large portion of the optimization results is close to 0. 
In the third phase, it is observed that a larger muscle tension is estimated by the optimization than the EMG signal.

%% file: contents/discussions.tex
\section{Discussions\label{sect:discussions}}
\subsection{Ground Reaction Force Estimation}
Tables \ref{tab:rrsme_id_walk} and \ref{tab:rrsme_id_run} show the root mean square error (RMSE) and related root mean square error (rRMSE) between the estimated and ground truth values, which are defined as follows \cite{oh2013prediction}:
\begin{align}
    {\rm RMSE} &= \sqrt{\frac{\sum^{n}_{i=1}(f_i-\hat{f}_i)^2}{n}}, \ \mbox{and}
    \\
    {\rm rRMSE} &= \frac{{\rm RMSE} \times 100}{1/2[\textit{diff}(f_i)+\textit{diff}(\hat{f}_i)]}
\end{align}
where $n$ is the number of data, $\textit{diff}(f)=\max(f)-\min(f)$, $f$ is the GRF estimated from the ID, and $\hat{f}$ is the ground truth measured by the force plate. 
These values were calculated only when the subject strode on each force plate.

\begin{table}[]
    \centering
    \caption{RMSE and rRMSE between estimated and measured contact force in the walking motion.}
    \label{tab:rrsme_id_walk}
    \begin{tabular}{|c|c|c|c|}
    \hline
        FP & x:RMSE(rRMSE) & y:RMSE(rRMSE) & z:RMSE(rRMSE)     \\ \hline
        1st         & 31.20N (11\%)  & 8.67N (8\%)   & 65.59N (9\%)   \\ \hline
        2nd         & 37.81N (14\%)  & 20.09N (18\%)  & 107.39N (14\%) \\ \hline
        3rd         & 39.77N (14\%)  & 11.58N (10\%)  & 97.84N (13\%)  \\ \hline
    \end{tabular}
\end{table}

\begin{table}[]
    \centering
    \caption{RMSE and rRMSE between estimated and measured contact force in the running motion.}
    \label{tab:rrsme_id_run}
    \begin{tabular}{|c|c|c|c|}
    \hline
        FP & x:RMSE(rRMSE) & y:RMSE(rRMSE) & z:RMSE(rRMSE)     \\ \hline
        1st         & 48.50N (13\%)  & 11.69N (10\%)   & 96.75N (7\%)   \\ \hline
        2nd         & 46.93N (12\%)  & 33.95N (29\%)  & 178.95N (12\%) \\ \hline
        3rd         & 76.63N (20\%)  & 24.83N (21\%)  & 222.23N (15\%)  \\ \hline
    \end{tabular}
\end{table}

\begin{enumerate}
    \item Walking motion result: 
    In Table \ref{tab:rrsme_id_walk}, the rRMSE for the walking motion was approximately 12\% in every axes.
From Fig. \ref{fig:result_id_plot_walk}, we observe that the error mainly occurs in the landing and pre-swing phases, when the foot or prosthesis is landing or departing from the ground.
Meanwhile, stable results are observed in the mid-phases. 
We consider that one of the reasons is the limitation of the contact point setting.
In the actual motion, the prosthesis has a rolling contact with the ground; however, it was simplified by a finite number of contacts as shown in \figref{fig:contact_point_setting}.

    \item Running motion result:
    As seen in Table \ref{tab:rrsme_id_run}, a relatively large rRMSE is observed compared to the walking motion, approximately 10\% to 20\% in every axes.
Similar to the walking motion, the error largely occurs during the landing and pre-swing phases. 
However, in contrast to the walking motion, we found that there were fluctuations in the x- and y-axes of the estimated GRF.
We consider that this is because the motion itself is faster than the walking motion, including a possibility that the motion data itself contains more noise.
\end{enumerate} 

Despite some limitations, the result of the ID is in good agreement with the measured ground truth with 12\% rRMSE. 
These results show the validity of the ID calculation of the hybrid-link system.

\begin{table}[tbp]
\centering
\caption{Root Mean Square Error between the estimations by the optimization and the EMG signal for each single phase.}
\label{tab:rmse_muscletension}
\begin{tabular}{|c|c|c|c|}
\hline
RMS Error{[}\%{]}            & 1st (Right) & 2nd (Left) & 3rd (Right) \\ \hline
Biceps Femoris Caput Longum & 7.4843       & 10.5783     & 10.4043      \\ \hline
Semitendinosus              & 8.8415       & 10.4481      & 12.2633       \\ \hline
\end{tabular}
\end{table}

\subsection{Muscle Force Estimation}
Table \ref{tab:rmse_muscletension} shows the root mean square error between the two estimations, divided by $\max(f_{\max i})$ of each muscle.
%($f_{\max i}$ is introduced in \eqref{fimax}). The result is shown in Table \ref{tab:rmse_muscletension}.
This result is calculated only for each single phase.
% Since $f_{\max i}$ is variable with muscle length, we need to choose the maximum value as a $\max(f_{\max i})$.
We consider that there are mainly two reasons in the error.
Firstly, there was a limitation on the position to attach the EMG sensors in this experiment. 
Usually, EMG sensor is attached on the middle of a muscle to obtain accurate result.
We consider the location of EMG sensor attached near the end part would degrade the accuracy. 
Secondly, we consider that there was a limitation of the optimization method itself. 
In Figs. \ref{fig:bicepsfemoriscaputlongum} and \ref{fig:semitendinosus}, continuous zero values are sometimes observed from the result. 
For a limitation of the muscle force optimization, the system cannot calculate the muscle force when the joint is in static movement such as the single support phase. 
Therefore, as a reason for zero values, we consider that the related joint was rarely moving in those moments. 
EMG signals are usually used for compensation of this problem, included in the optimization \eqref{eq:ms_optimization}.
Therefore, it is expected that a better estimation result will be obtained using EMG signal in the optimization.

Although there are limitations, Table \ref{tab:rmse_muscletension} shows low values for RMSE, especially for the first stance phase with the prosthetic leg. 
From these results, we consider that the proposed musculoskeletal model with hybrid-link system can provide an estimation of muscle force, explicitly considering the deformation of the flexible prosthesis and muscle amputation.

Although we set $\bm{W}_2 = \bm{O}$ for comparison, appropriate setting of $\bm{W}_2$ would improve the estimation of muscle forces, using EMG signals as reference.
This is one of our future works.

%% file: contents/conclusion.tex
\section{Conclusion} \label{section:conclusion}
In this study, we introduced a framework for the dynamical analysis of an athlete motion wearing a leaf-spring prosthesis, based on the inverse dynamics of a soft-rigid hybrid multi-link system.
The obtained results can be summarized as follows:
\begin{enumerate}
    \item We built a hybrid-link model that integrated human skeleton system with a leaf-spring prosthesis, in which the flexible deformation of the prosthesis was represented by the PCS model developed in soft robotics research field.
    We implemented the inverse kinematics of the hybrid-link system and applied it to the motion reconstruction from the motion captured data.
    \item Using the QP-based inverse dynamics, we can estimate the joint torques and ground reaction force during a motion, explicitly considering the interaction force between human limb and prosthesis leg.
    Comparing the rRMSE between the measured force and estimated force, the results showed that the error was low in the mid phase but high in the landing and pre-swing phases.
    Despite limitations, we showed that the rRMSE was approximately 12\% in the walking motion.
    \item We can estimate muscle tension force by the optimization using the result obtained from the inverse dynamics.
    We built a muscle model to replicate the muscle amputation and estimated the muscle force during a running motion.
    We quantitatively compared it with another estimation by EMG signals and showed that the RMSE between two values shows around 10\% error for the maximum muscle tension. 
\end{enumerate}

A known limitation in this study is that all results were made for only one subject.
This was because it was hard to find a subject who wears a leaf-spring prosthesis playing sports and can participate in an experiment.
Future experiments with much more subjects will provide more reliable results. 
We believe that this study contributes to an important step for biomechanical analysis in Para-sport field as a practical application of the traditional robotics and latest soft robotics research.

%% file: contents/apdx_msforce_calc.tex
\section{Muscle Tension Force Calculation using EMG Signal\label{apdx:msforce_calc}}
In muscle model proposed by Hill \cite{hill1938heat}, the muscle is modeled by using the muscle length and velocity. 
To complement the  intrinsic muscle stiffness which is not properly represented in Hill-type muscle model, Stroeve \cite{stroeve1999impedance} proposed neuro-musculo-skeletal model, using the muscle model as a simplification of \cite{winters1985analysis}. 
According to \cite{stroeve1999impedance}, an arbitrary muscle tension force $f_i$ is given by
\begin{align}
    \label{eq:hill_stroeve_muscle_tension}
    f_i = -{a_i}F_l(l_i)F_v({\dot{l}_i})F_{\max i}
\end{align}
where ${a_i}$ is muscle activity; 
$l_i$ and ${\dot{l}_i}$ are muscle length and velocity, respectively;
$F_{\max i}$ is maximum isometric force of muscle $i$;
$F_l(l_i)$ is Gaussian force-length relation; and
$F_v({\dot{l}_i})$ is Hill-type force-velocity relation.

Muscle activity at current state $a_i(t)$ is updated from the EMG signal as follows:
\begin{align}
    \label{eq:emg_mf}
    \dot{a}_i(t) &= \frac{u_i-a_i(t-1)}{\tau^{*}} 
    \\
    \tau^{*} &=
    \begin{cases}
        \tau_{ac} & u_i \geq a_i(t-1) \\
        \tau_{da} & u_i < a_i(t-1)
    \end{cases}
    \\
    u_i &= \frac{u_{\rm{EMG}}}{u_{\rm{MVC}}} 
\end{align}
where $\tau_{ac}$ is an activation time constant; 
$\tau_{da}$ is a deactivation time constant; and
$u_i$ is the neural input obtained from EMG signal $u_{\rm{EMG}}$ and Maximum Voluntary Contraction (MVC) of corresponding muscle $u_{\rm{MVC}}$.
Finally, we can compute the muscle tension $f_i$ from the EMG signal.

%% file: contents/apdx_hybrid_link_inertia.tex
\section{Inertia Matrix of Hybrid-link System\label{apdx:hybrid_link_inertia}}
   The inertia matrix $\bm{M}(\bm{q})$ in Eq. (29) or (30) can be derived in a manner similar to that of a rigid-body multi-link system.
   First, the kinetic energy of a single rigid-link is represented as
   \begin{align}
       T_{R,i} = \frac{1}{2} \bm{\eta}_i^T \mathcal{M}_{R,i} \bm{\eta}_i,
       \\
       \mathcal{M}_{R,i}
       :=
       \begin{bmatrix}
           \bm{I}_{G,i} & \bm{O} \\
           \bm{O} & m_i \bm{E}
       \end{bmatrix}
   \end{align}
   where $\bm{I}_{G,i}$ is the $3 \times 3$ inertia matrix about the center of gravity of a rigid link.
   Then, $T_{R,i}$ can be rewritten as
   \begin{align}
        \label{eq:kinetic_energy_rigid}
       T_{R,i} =
       \frac{1}{2} \bm{\psi}^T \bm{J}_{R,i}^T \mathcal{M}_{R,i} \bm{J}_{R,i} \bm{\psi}
   \end{align}
   using $\bm{\eta}_i = \bm{J}_{R,i} \bm{\psi}$ with the Jacobian matrix $\bm{J}_{R,i}$.
   
   Then, the kinetic energy of a segment in the PCS model can be represented by the same formulation as
   \begin{align}
        \label{eq:kinetic_energy_pcs}
       T_{S,i} =
       \frac{1}{2} \bm{\psi}^T \bm{J}_{S,i}^T \mathcal{M}_{S,i} \bm{J}_{S,i} \bm{\psi}
   \end{align}
   because of the similarity of the mathematical structure between the rigid-body and PCS model although the detailed contents of $\mathcal{M}_{S,i}$ is more complicated than $\mathcal{M}_{R,i}$.

   From Eqs. (\ref{eq:kinetic_energy_rigid}) and (\ref{eq:kinetic_energy_pcs}), the total kinetic energy of a hybrid-link system can be represented as
   \begin{align}
       T = \sum_{i} T_{R,i} + \sum_{i} T_{S,i}
       =
       \frac{1}{2} \bm{\psi}^T \left(
       \sum_i\bm{J}_{R,i}^T \mathcal{M}_{R,i} \bm{J}_{R,i}
       +
       \sum_i\bm{J}_{S,i}^T \mathcal{M}_{S,i} \bm{J}_{S,i}
       \right)
       \bm{\psi}
   \end{align}
   Therefore, the total inertia matrix of $\bm{M}(\bm{q})$ is derived as
   \begin{align}
        \label{eq:inertia_matrix_hybrid}
       \bm{M}(\bm{q})
       =
       \sum_i\bm{J}_{R,i}^T \mathcal{M}_{R,i} \bm{J}_{R,i}
       +
       \sum_i\bm{J}_{S,i}^T \mathcal{M}_{S,i} \bm{J}_{S,i}
   \end{align}
   Each block in $\bm{M}(\bm{q})$ such as $\bm{M}_{0}$ and $\bm{M}_{R0}$ is derived from Eq. (\ref{eq:inertia_matrix_hybrid}).
   In practice, however, it is more efficient to numerically compute than to analytically calculate those matrices, e.g., by the unit vector method.
   Or, the articulated body algorithm, which has been proposed also for the PCS model \cite{renda2018discrete}, does not explicitly compute those matrices but directly compute the acceleration.